\documentclass[acmtog, nonacm]{acmart}
\renewcommand\footnotetextcopyrightpermission[1]{} 
\usepackage{booktabs} 

\usepackage[colorinlistoftodos]{todonotes}

\usepackage[ruled]{algorithm2e} 

\SetAlFnt{\small}
\SetAlCapFnt{\small}
\SetAlCapNameFnt{\small}
\SetAlCapHSkip{0pt}

\acmYear{2020}
\acmMonth{8}

\setcopyright{none}

\acmDOI{}

\begin{document}

\title{Learning Illumination from Diverse Portraits}

\author{Chloe LeGendre}
\affiliation{%
  \institution{Google Research}
 }
\email{chlobot@google.com}

\author{Wan-Chun Ma}

\author{Rohit Pandey}

\author{Sean Fanello}

\author{Christoph Rhemann}

\author{Jason Dourgarian}

\author{Jay Busch}
\affiliation{%
  \institution{Google}
}

\author{Paul Debevec}
\affiliation{%
  \institution{Google Research}
}

\renewcommand\shortauthors{C. LeGendre et al.}

\begin{teaserfigure}
\centering
\vspace{-3pt}
\begin{tabular}{@{}c@{\quad}c@{\quad}c@{\quad}c@{}}
			\includegraphics[height=1.47in]{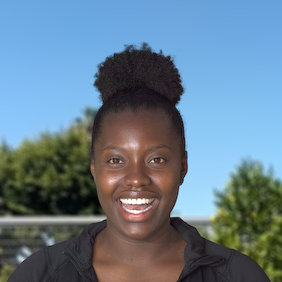} &
			\includegraphics[height=1.47in]{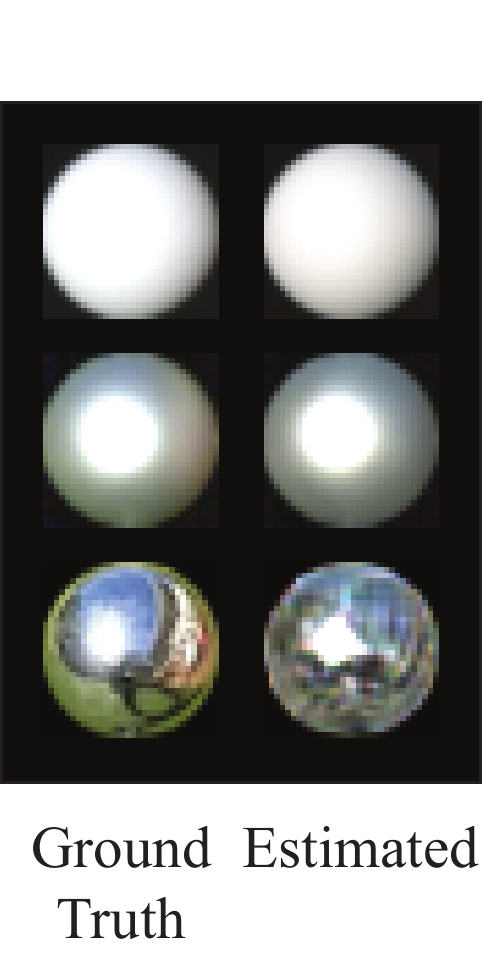} &
			\includegraphics[height=1.47in]{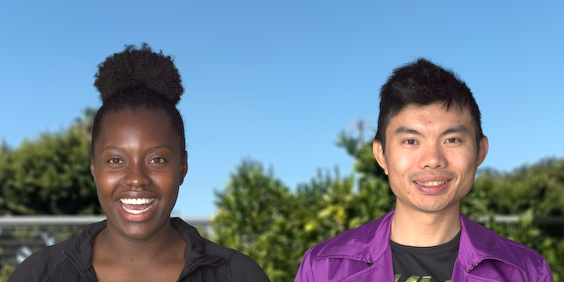} &
			\includegraphics[height=1.47in]{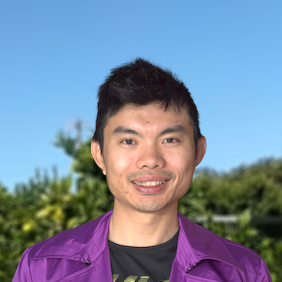} \\
			\small{\textbf{(a)} Input portrait} & \small{\textbf{(b)} Lighting} & \small{\textbf{(c)} Original and novel subjects lit with estimated lighting} & \small{\textbf{(d)} Novel subject lit by GT lighting} \\
\end{tabular}
\vspace{-5pt}
\caption{Our network estimates HDR omnidirectional lighting from an LDR portrait image. \textbf{(a)} Input portrait image generated using a photographed reflectance basis. \textbf{(b)} Ground truth and estimated lighting, shown on diffuse, glossy, and mirror spheres. \textbf{(c)} Original and novel subjects lit consistently by the estimated lighting, using image-based relighting. \textbf{(d)} The novel subject lit with the original subject's ground truth lighting. For both subjects, the appearance under the estimated lighting closely matches the appearance under the original lighting.
}
\label{fig:teaser}
\end{teaserfigure}

\begin{abstract}
We present a learning-based technique for estimating high dynamic range (HDR), omnidirectional illumination from a single low dynamic range (LDR) portrait image captured under arbitrary indoor or outdoor lighting conditions. We train our model using portrait photos paired with their ground truth environmental illumination.  We generate a rich set of such photos by using a light stage to record the reflectance field and alpha matte of 70 diverse subjects in various expressions.  We then relight the subjects using image-based relighting with a database of one million HDR lighting environments, compositing the relit subjects onto paired high-resolution background imagery recorded during the lighting acquisition. We train the lighting estimation model using rendering-based loss functions and add a multi-scale adversarial loss to estimate plausible high frequency lighting detail. We show that our technique outperforms the state-of-the-art technique for portrait-based lighting estimation, and we also show that our method reliably handles the inherent ambiguity between overall lighting strength and surface albedo, recovering a similar scale of illumination for subjects with diverse skin tones. We demonstrate that our method allows virtual objects and digital characters to be added to a portrait photograph with consistent illumination.  Our lighting inference runs in real-time on a smartphone, enabling realistic rendering and compositing of virtual objects into live video for augmented reality applications.
\end{abstract}

\maketitle
\thispagestyle{plain}

\section{Introduction}
In both portrait photography and film production, lighting greatly influences the look and feel of a given shot. Photographers and cinematographers dramatically light their subjects to communicate a particular aesthetic sensibility and emotional tone. While films using visual effects techniques often blend recorded camera footage with computer-generated, rendered content, the realism of such composites depends on the consistency between the real-world lighting and that used to render the virtual content. Thus, visual effects practitioners work painstakingly to capture and reproduce real-world illumination inside \textit{virtual} sets. Debevec \shortcite{debevec:1998:rendering} introduced one such technique for real-world lighting capture, recording the color and intensity of omnidirectional illumination by photographing a mirror sphere using multiple exposures. This produced an HDR "image-based lighting" (IBL) environment \cite{debevec:2006:ibl}, used for realistically rendering virtual content into real-world photographs.

Augmented reality (AR) shares with post-production visual effects the goal of realistically blending virtual content and real-world imagery. Face-based AR applications are ubiquitous, with widespread adoption in both social media and video conferencing applications. However, in real-time AR, lighting measurements from specialized capture hardware are unavailable, as acquisition is impractical for casual mobile phone or headset users. Similarly, in visual effects, on-set lighting measurements are not always available, yet lighting artists must still reason about illumination using cues in the scene. If the footage includes faces, their task is somewhat less challenging, as faces include a diversity of surface normals and reflect light somewhat predictably. 

Prior work has leveraged the strong geometry and reflectance priors from faces to solve for lighting from portraits. In the years since Marschner and Greenberg \shortcite{marschner:1997:inverse} introduced portrait "inverse lighting," most such techniques \cite{shim:2012:faces, knorr:2014:real, tewari:2017:mofa, sengupta:2018:sfsnet, tewari:2018:FaceModel, egger:2018:occlusion, zhou:2018:label, kemelmacher:2010:3d, shu:2017:neural} have sought to recover both facial geometry and a low frequency approximation of distant scene lighting, usually represented using up to a 2\textsuperscript{nd} order spherical harmonic (SH) basis. The justification for this approximation is that skin reflectance is predominantly diffuse (Lambertian) and thus acts as a low-pass filter on the incident illumination. For diffuse materials, irradiance indeed lies very close to a 9D subspace well-represented by this basis \cite{ramamoorthi:2001:relationship, basri:2003:lambertian}. 

However, to the skilled portrait observer, the lighting at capture-time reveals itself not only through the skin's diffuse reflection, but also through the directions and extent of cast shadows and the intensity and locations of specular highlights. Inspired by these cues, we train a neural network to perform inverse lighting from portraits, estimating omnidirectional HDR illumination without assuming any specific skin reflectance model. Our technique yields higher frequency lighting that can be used to convincingly render novel subjects into real-world portraits, with applications in both visual effects and AR when off-line lighting measurements are unavailable. Furthermore, our lighting inference runs in real-time on a smartphone, enabling such applications.

\vspace{-7pt}
\begin{figure}[ht]
\centering
\includegraphics[width=2.7in]{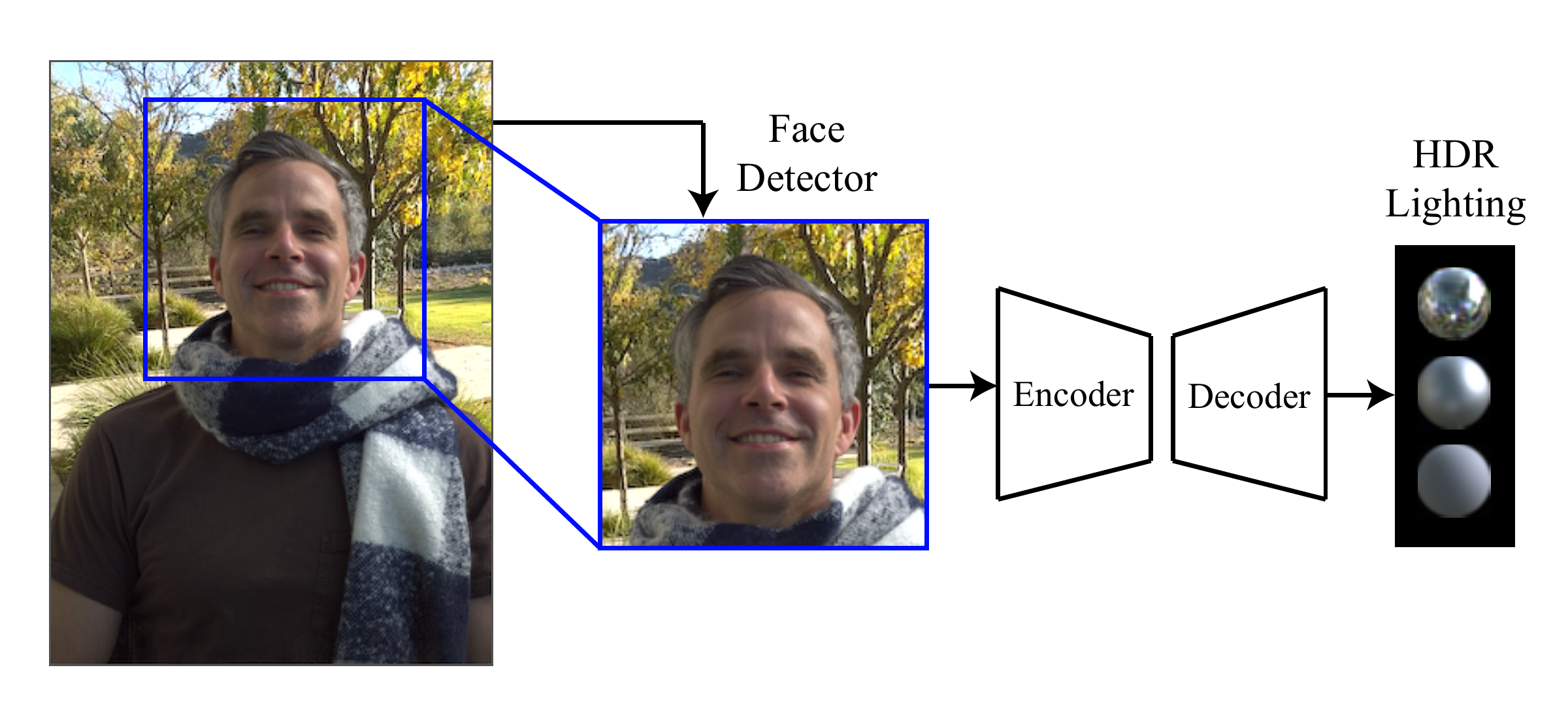}
\vspace{-10pt}
\caption{We train a convolutional neural network to regress from a face-cropped input image to omnidirectional, HDR illumination.}
\label{fig:facelight_overview}
\end{figure}
\vspace{-5pt}

We train our lighting estimation model in a supervised manner using a dataset of portraits and their corresponding ground truth illumination. To generate this dataset, we photograph 70 diverse subjects in a light stage system as illuminated by 331 directional light sources forming a basis on a sphere, such that the captured subject can be relit to appear as they would in any scene with image-based relighting \cite{debevec:2000:acquiring}. Although a few databases of real-world lighting environments captured using traditional HDR panoramic photography techniques are publicly available, e.g. the Laval indoor and outdoor datasets with 2,000 and 12,000 scenes respectively \cite{gardner:2017:indoor, lalonde:2014:collections}, we extend the LDR data collection technique of LeGendre et al. \shortcite{legendre:2019:deeplight} to instead capture on the order of 1 million indoor and outdoor lighting environments, promoting them to HDR via a novel non-negative least squares solver formulation before using them for relighting. 

A few recent works have similarly sought to recover illumination from portraits without relying on a low-frequency lighting basis, including the deep learning methods of Sun et al. \shortcite{sun:2019:single} for arbitrary scenes and Calian et al. \shortcite{calian:2018:faces} for outdoor scenes containing the sun. We show that our method out-performs both of these methods, and generalizes to arbitrary indoor or outdoor scenes.

Any attempt at lighting estimation is complicated by the inherent ambiguity between surface reflectance (albedo) and light source strength \cite{belhumeur:1999:bas}. Stated otherwise, a pixel's shading is rendered unchanged if its albedo is halved while light source intensity doubles. Statistical priors for facial albedo have been leveraged to resolve this ambiguity \cite{egger:2018:occlusion, tewari:2017:mofa, calian:2018:faces}, but, to the best of our knowledge, we are the first to explicitly evaluate the performance of our model on a wide variety of subjects with different skin tones. In contrast, Sun et al. \shortcite{sun:2019:single} report lighting accuracy with a scale-invariant metric, while Calian et al. \shortcite{calian:2018:faces} show visual results for synthetically rendered and photographed faces where the subjects are predominantly light in skin tone. We show that for a given lighting condition, our model can recover lighting at a similar scale for a variety of diverse subjects.

In summary, our contributions are the following:
\begin{itemize}
\item A deep learning method to estimate HDR illumination from LDR images of faces in both indoor and outdoor scenes. Our technique outperforms the previous state-of-the-art. 
\item A first-of-its-kind analysis that shows that our HDR lighting estimation technique reliably handles the ambiguity between light source strength and surface albedo, recovering similar illumination for subjects with diverse skin tones. 
\end{itemize}

\section{Related Work}
In this section we summarize work related to lighting capture, inverse rendering from faces, and the related topics of portrait relighting and unconstrained lighting estimation.

\paragraph{Lighting measurement techniques.} After Debevec \shortcite{debevec:1998:rendering} introduced image-based lighting from high dynamic range panoramas, subsequent work proposed more general acquisition techniques including recording the extreme dynamic range of sunny daylight with a fisheye lens \cite{stumpfel:2004:direct} and recording HDR video with a mirror sphere \cite{unger:2006:dsl, waese:2002:art}. Debevec et al. \shortcite{debevec:2012:single} and Reinhard et al. \shortcite{reinhard:2010:high} presented more practical techniques to recover the full dynamic range of daylight by augmenting the typical mirror sphere capture with simultaneous photography of a diffuse, gray sphere that allowed for saturated light source intensity recovery.  We extend these techniques to promote one million real-world, clipped panoramas to HDR.

\paragraph{Inverse Rendering.} The joint recovery of scene geometry, material reflectance, and illumination given only an image, thereby inverting the image formation or rendering process, is a long-studied problem in computer vision \cite{yu:1999:inverse, ramamoorthi:2001:signal, lombardi:2016:reflectance}. Similarly, the topic of "intrinsic image" decomposition has received considerable attention, recovering shading and reflectance, rather than geometry and illumination \cite{land:1971:lightness, barrow:1978:recovering}. "Shape from Shading" methods aim to recover geometry under known illumination \cite{horn:1970:shape}, while another variant jointly recovers "Shape, Illumination, and Reflectance from Shading" \cite{barron:2014:shape}. 

Recently, significant progress has been made in the domain of inverse rendering from portrait images or videos, with the goal of recovering a 3D face model with illumination and/or reflectance \cite{kemelmacher:2010:3d, egger:2018:occlusion, tewari:2018:FaceModel, tewari:2017:mofa, sengupta:2018:sfsnet, yamaguchi:2018, tran:2019:towards, tran:2019:learning}. Many of these techniques rely on geometry estimation via fitting or learning a 3D morphable model \cite{blanz:1999:morphable}, and they model skin reflectance as Lambertian and scene illumination using a low-frequency 2\textsuperscript{nd} order SH basis. In contrast, our goal is to recover higher frequency illumination useful for rendering virtual objects with diverse reflectance characteristics beyond Lambertian.

\paragraph{Inverse Lighting from Faces.} Marschner and Greenberg \shortcite{marschner:1997:inverse} introduced the problem of "inverse lighting," estimating the directional distribution and intensity of incident illumination falling on a rigid object with measured geometry and reflectance, demonstrating lighting estimation from portraits as one such example. With the appropriate lighting basis and reflectance assumption, the problem was reduced to inverting a linear system of equations. The linearity of light transport was similarly leveraged in follow-up work to estimate lighting from faces \cite{shim:2012:faces, shahlaei:2015:realistic}, including for real-time AR \cite{knorr:2014:real}, but these approaches estimated either a small number of point light sources or again used a low frequency 2\textsuperscript{nd} order SH lighting basis. Specular reflections from the eyes of portrait subjects have been leveraged to estimate higher frequency illumination, but as the reflections of bright light sources are likely to be clipped, the recovery of the full dynamic range of natural illumination is challenging to recover from a single exposure image \cite{nishino:2004:eyes}. 

Several new deep learning techniques for inverse lighting from faces have been proposed. Zhou et al. \shortcite{zhou:2018:label} estimated 2\textsuperscript{nd} order SH illumination from portraits. For higher frequency lighting estimates, Yi et al. \shortcite{yi:2018:faces} recovered illumination by first estimating specular highlights and ray-tracing them into a panorama of lighting directions. However, this model produced HDR IBL maps that are mostly empty (black), with only dominant light source colors and intensities represented. In contrast, we estimate plausible omnidirectional illumination. Calian et al. \shortcite{calian:2018:faces} trained an autoencoder on a large database of outdoor panoramas to estimate lighting from LDR portraits captured outdoors, combining classical inverse lighting and deep learning. While this method produced impressive results for outdoor scenes with natural illumination, it is not applicable to indoor scenes or outdoor scenes containing other sources of illumination. Our model, in contrast, generalizes to arbitrary settings. Critically, neither Yi et al. \shortcite{yi:2018:faces} nor Calian et al. \shortcite{calian:2018:faces} evaluated model performance on subjects with diverse skin tones, which we feel is an important variation axis for lighting estimation error analysis. Both works presented qualitative results only for photographed subjects and rendered computer-generated models with fair skin.

\paragraph{Portrait Relighting.} Marchner and Greenberg \shortcite{marschner:1997:inverse} also proposed \textit{portrait relighting} and \textit{portrait lighting transfer}, showing that the lighting from one portrait subject could be used to approximately relight another subject, such that the two could be convincingly composited together into one photograph. Recent works solved this problem either with a mass transport \cite{shu:2017:portrait} or deep learning \cite{sun:2019:single, zhou:2019:deep} approach. Sun et al. \shortcite{sun:2019:single} estimated illumination while training a portrait relighting network. Lighting estimates from this technique proved superior compared with two other state-of-the-art methods \cite{sengupta:2018:sfsnet, barron:2014:shape}. Similarly to Sun et al. \shortcite{sun:2019:single}, we generate photo-realistic, synthetic training data using a set of reflectance basis images captured in an omnidirectional lighting system, or light stage, relying on the technique of \textit{image-based relighting} \cite{nimeroff:1995:efficient, debevec:2000:acquiring} to synthesize portraits lit by novel sources of illumination. However, in contrast to Sun et al. \shortcite{sun:2019:single}, we extend a recent environmental lighting capture technique \cite{legendre:2019:deeplight} to expand the number of lighting environments used for training data, employ a set of loss functions designed specifically for lighting estimation, and use a lightweight model to achieve lighting inference at real-time frame rates on a mobile device. Even when trained on the same dataset, we show that our lighting estimation model outperforms that of Sun et al. \shortcite{sun:2019:single}, the previous state-of-the-art for lighting estimation from portraits.

\paragraph{Lighting Estimation.}
Given the prominence of virtual object compositing in both visual effects and AR, it is unsurprising that lighting estimation from arbitrary scenes (not from portraits) is also an active research area. Several works have sought to recover outdoor, natural illumination from an unconstrained input image \cite{lalonde:2009:estimating, lalonde:2014:collections, hold:2017:deep, hold:2019:deepsky, zhang:2019:allweather}. Several deep learning based methods have recently tackled indoor lighting estimation from unconstrained images \cite{song:2019:neural, garon:2019:fast, gardner:2017:indoor}. Cheng et al. \shortcite{cheng:2018:learning} estimated lighting using a deep learning technique given two opposing views of a panorama. For AR applications, LeGendre at al. \shortcite{legendre:2019:deeplight} captured millions of LDR images of three diffuse, glossy, and mirror reference spheres as they appeared in arbitrary indoor and outdoor scenes, using this dataset to train a model to regress to omnidirectional HDR lighting from an unconstrained image. We leverage this lighting data collection technique but extend it to explicitly promote the captured data to HDR so that it can be used for image-based relighting, required for generating our synthetic portraits. LeGendre et al. \shortcite{legendre:2019:deeplight} trained their model using a combination of rendering-based and adversarial losses, which we extend to the multi-scale domain for superior performance. 
\section{Method}

\subsection{Training Data Acquisition and Processing}
\label{sec:data}
To train a model to estimate lighting from portrait photographs in a supervised manner, we require many portraits labeled with ground truth illumination. Since no such real-world, dataset exists, we synthesize portraits using the data-driven technique of image-based relighting, shown by Debevec et al. \shortcite{debevec:2000:acquiring} to produce photo-realistic relighting results for human faces, appropriately capturing complex light transport phenomena for human skin and hair e.g. sub-surface and asperity scattering and Fresnel reflections. Noting the difficulty of generating labeled imagery for the problem of inverse lighting from faces, many prior works have instead relied on renderings of 3D models of faces \cite{calian:2018:faces, yi:2018:faces, zhou:2018:label}, which often fail to represent these complex phenomena. 

\begin{figure}[ht]
\vspace{-2pt}
\centering
\includegraphics[width=2.8in]{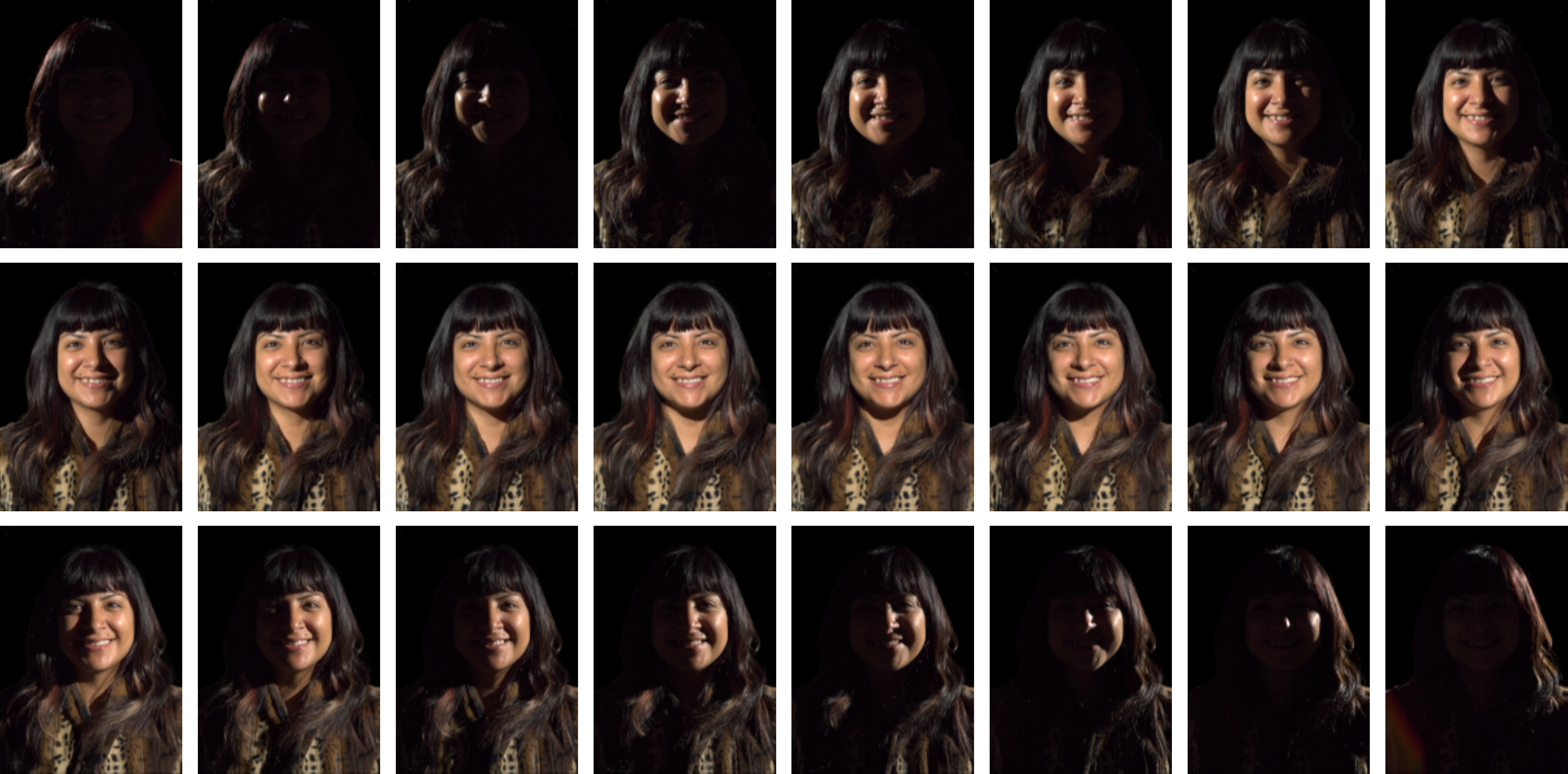}
\vspace{-5pt}
\caption{"One-Light-at-a-Time" images: 24 of the 331 lighting directions.}
\vspace{-10pt}
\label{fig:olats}
\end{figure}

\begin{figure}[ht]
\vspace{-2pt}
\centering
\includegraphics[width=2.5in]{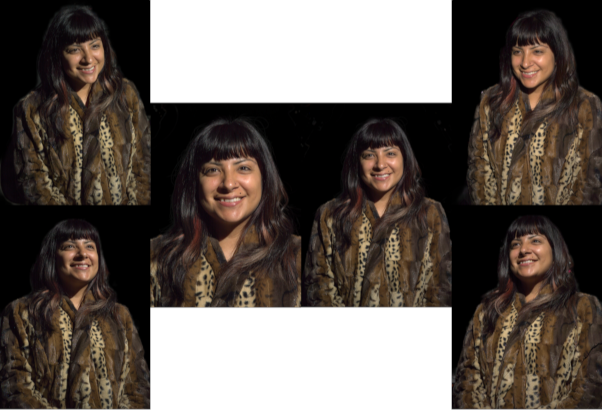}
\vspace{-5pt}
\caption{
Our six camera viewpoints for an example lighting direction.}
\vspace{-5pt}
\label{fig:camera_views}
\end{figure}

\paragraph{Reflectance Field Capture.} Debevec et al. \shortcite{debevec:2000:acquiring} introduced the 4D \textit{reflectance field} $R(\theta,\phi, x, y)$ to denote a subject lit from any lighting direction $(\theta,\phi)$ for image pixels $(x, y)$ and showed that taking the dot product of this reflectance field with an HDR lighting environment similarly parameterized by $(\theta,\phi)$ relights the subject to appear as they would in that scene. To photograph a subject's reflectance field, we use a computer-controllable sphere of 331 white LED light sources, similar to that of Sun et al. \shortcite{sun:2019:single}, with lights spaced $12^{\circ}$ apart at the equator. The reflectance field is formed from a set of reflectance basis images (see Fig. \ref{fig:olats}), photographing the subject as each of the directional LED light sources is individually turned on one-at-a-time within the spherical rig. We capture these "One-Light-At-a-Time" (OLAT) images for multiple camera viewpoints, shown in Fig. \ref{fig:camera_views}. In total we capture $331$ OLAT images for each subject using six color Ximea machine vision cameras with 12 megapixel resolution, placed $1.7$ meters from the subject. The cameras are positioned roughly in front of the subject, with five cameras with 35 mm lenses capturing the upper body of the subject from different angles, and one additional camera with a 50 mm lens capturing a close-up image of the face with tighter framing. 

\begin{figure}[ht]
\vspace{-3pt}
\centering
\includegraphics[width=\linewidth]{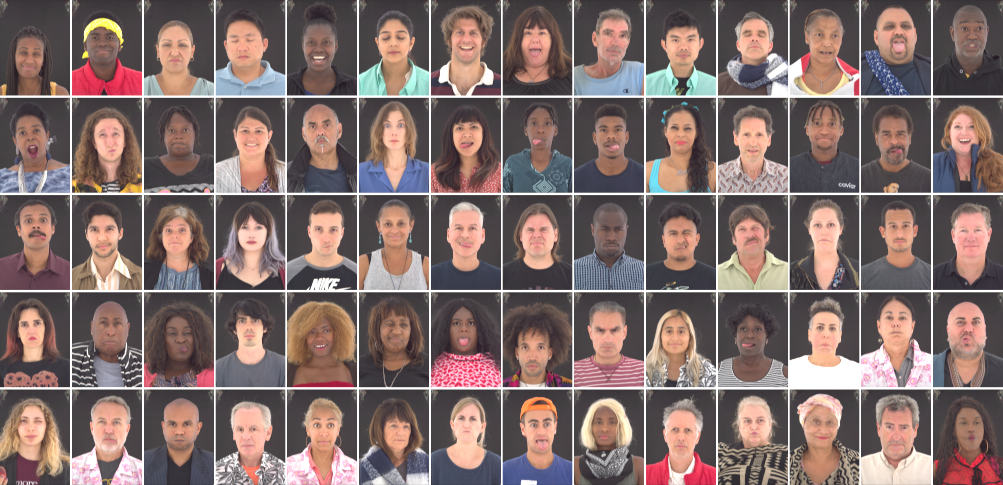}
\vspace{-13pt}
\caption{
Representative portraits of the 70 recorded subjects.}
\label{fig:subjects}
\vspace{-12pt}
\end{figure}

We capture reflectance fields for 70 diverse subjects, each performing nine different facial expressions and wearing different accessories, yielding about 630 sets of OLAT sequences from six different camera viewpoints, for a total of 3780 unique OLAT sequences. In addition to age and gender diversity, we were careful to photograph subjects spanning a wide range of skin tones, as seen in Fig. \ref{fig:subjects}. 

Since acquiring a full OLAT sequence for a subject takes six seconds, there can be subject motion over the sequence. We therefore employ an optical flow technique \cite{anderson:2016:jump} to align the images, interspersing at every 11\textsuperscript{th} OLAT frame one extra "tracking" frame with even, consistent illumination to ensure the brightness constancy constraint for optical flow is met, as in Wenger et al. \shortcite{wenger:2005:performance}. This step preserves the sharpness of image features when performing the relighting operation, which linearly combines aligned OLAT images.

\begin{figure}[ht]
	\centerline{
		\begin{tabular}{@{ }c@{ }c@{ }c@{ }c} 
		    \includegraphics[width=0.75in]{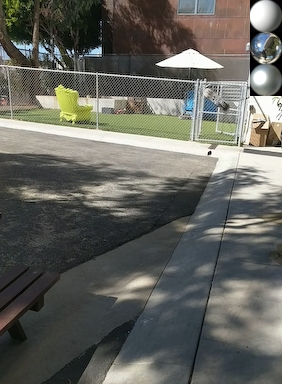} &
			\includegraphics[width=0.75in]{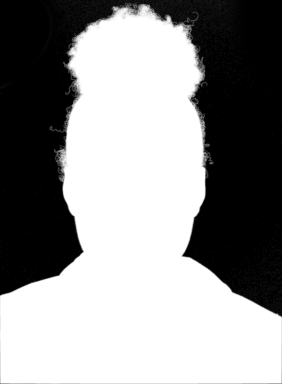} &
			\includegraphics[width=0.75in]{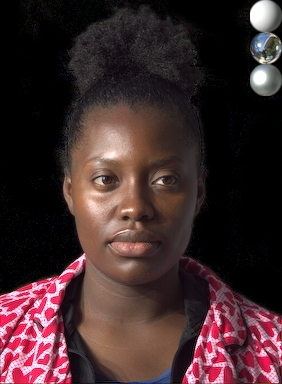} &
			\includegraphics[width=0.75in]{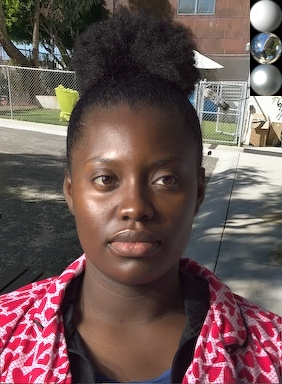} \\
			\includegraphics[width=0.75in]{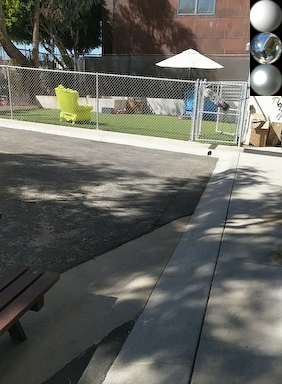} &
			\includegraphics[width=0.75in]{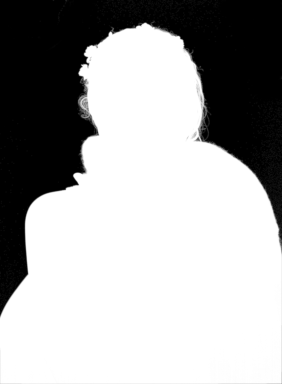} &
			\includegraphics[width=0.75in]{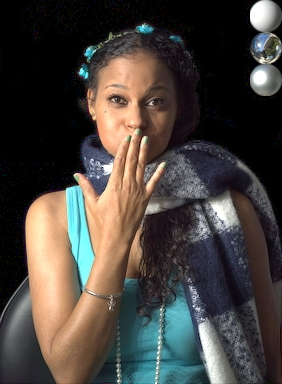} &
			\includegraphics[width=0.75in]{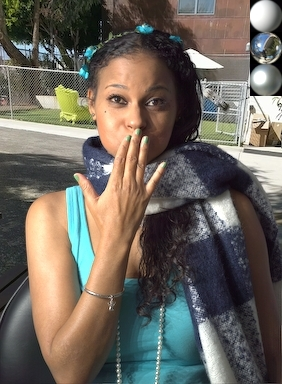} \\
			\small{\textbf{(a)} background} & \small{\textbf{(b)} alpha matte} & \small{\textbf{(c)} relit subject} & \small{\textbf{(d)} composited} \\
		\end{tabular}}
		\vspace{-5pt}
		\caption{\textbf{(a)} A background with paired HDR illumination, shown via the inset spheres (upper right). \textbf{(b)} Alpha matte from our system. \textbf{(c)} Subject relit with the illumination from \textbf{a}. \textbf{(d)} Subject relit and composited into \textbf{a.}}
		\label{fig:compositing}
		\vspace{-15pt}
\end{figure}

\paragraph{Alpha Matte Acquisition.} For the two frontal camera views, we also acquire images to compute an alpha matte for each subject, so we can composite them over novel backgrounds. We acquire a first image where the subject is unlit and a grey background material placed behind the subject is lit relatively evenly by six LED sources. We also photograph a clean plate of the background under the same lighting condition without the subject in the scene, such that the alpha matte can be computed by dividing the first image by the clean plate, as in Debevec et al. \shortcite{debevec:2002:lighting}. Although the work of Sun et al. \shortcite{sun:2019:single} uses a human segmentation algorithm to remove sections of the image corresponding to background elements (e.g. the light stage rig and lights visible within the camera view), we use our more accurate alpha matte for our frontal views (Fig. \ref{fig:compositing}b). For the remaining views, we compute an approximate segmentation using a method designed to handle the challenging task of segmenting hair from background imagery \cite{tkachenka:2019:real}. 

\begin{figure}[t]
\vspace{-3pt}
\centering
\includegraphics[width=\linewidth]{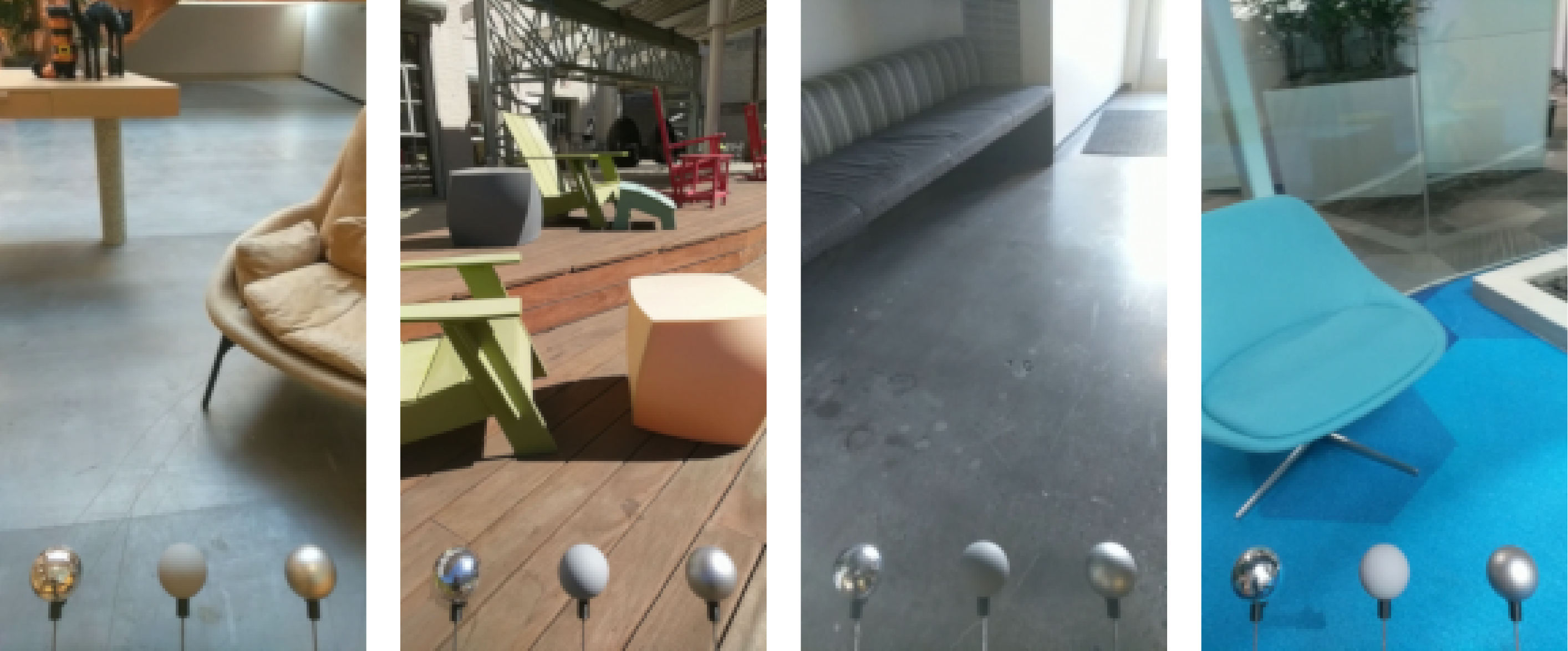}
\vspace{-13pt}
\caption{Background images with ground truth lighting recorded by diffuse, matte silver, and mirrored spheres as in LeGendre et al. \shortcite{legendre:2019:deeplight}.}
\label{fig:prober}
\vspace{-15pt}
\end{figure}

\paragraph{HDR Lighting Environment Capture.} To relight our subjects with photographed reflectance fields, we require a large database of HDR lighting environments, where no light sources are clipped. While there are a few such datasets containing on the order of thousands of indoor panoramas \cite{gardner:2017:indoor} or the upper hemisphere of outdoor panoramas \cite{lalonde:2014:collections}, deep learning models are typically enhanced with a greater volume of training data. Thus, we extend the video-rate capture technique of LeGendre et al. \shortcite{legendre:2019:deeplight} to collect on the order of 1 million indoor and outdoor lighting environments. This work captured background images augmented by a set of diffuse, matte silver, and mirrored reference spheres held in the lower part of the frame as in Fig. \ref{fig:prober}. These three spheres reveal different cues about the scene illumination. The mirror ball reflects omnidirectional high frequency lighting, but bright light sources will be clipped, altering both their intensity and color. The near-Lambertian BRDF of the diffuse ball, in contrast, acts as a low-pass filter on the incident illumination, capturing a blurred but relatively complete record of total scene irradiance. Without explicitly promoting these LDR sphere appearances to a record of HDR environmental illumination, LeGendre et al. \shortcite{legendre:2019:deeplight} regressed from the unconstrained background images to HDR lighting using an in-network, differentiable rendering step, predicting illumination to match the clipped, LDR ground truth sphere appearances. In contrast, we require a true HDR record of the scene illumination to use for relighting our subjects, so, unlike LeGendre et al. \shortcite{legendre:2019:deeplight}, we must explicitly promote the three sphere appearances into an estimate of their corresponding HDR lighting environment.

\paragraph{Promoting LDR Sphere Images to HDR Lighting.} Given captured images of the three reflective spheres, perhaps with clipped pixels, we wish to solve for HDR lighting that could have plausibly produced these three sphere appearances. We first record the reflectance field for the diffuse and matte silver spheres, again using the light stage system.  We convert their reflectance basis images into the same relative radiometric space, normalizing based on the incident light source color.  We then project the reflectance basis images into the mirror ball mapping \cite{reinhard:2010:high} (Lambert azimuthal equal-area projection), accumulating energy from the input images for each new lighting direction $(\theta,\phi)$ on a $32 \times 32$ image of a mirror sphere as in LeGendre et al. \shortcite{legendre:2019:deeplight}, forming the reflectance field $R(\theta,\phi, x, y)$, or, sliced into individual pixels, $R_{x,y}(\theta, \phi)$.

For lighting directions  $(\theta, \phi)$ in the captured mirror ball image \textit{without} clipping for color channel $c$, we recover the scene lighting $L_{c}(\theta, \phi)$ by simply scaling the mirror ball image pixel values by the inverse of the measured mirror ball reflectivity (82.7\%). For lighting directions $(\theta, \phi)$ \textit{with} clipped pixels in the original mirror ball image, we set the pixel values to $1.0$, scale this by the inverse of the measured reflectivity forming $L_{c}(\theta, \phi)$, and then subsequently solve for a residual missing lighting intensity $U_{c}(\theta, \phi)$ using a non-negative least squares solver formulation. Given an original image pixel value $p_{x,y,c,k}$ for BRDF index $k$ (e.g. diffuse or matte silver), and color channel $c$, and the measured reflectance field $R_{x, y, c, k}(\theta,\phi)$, due to the superposition principle of light, we can write: 

\vspace{-4pt}
\begin{equation}
p_{x,y,c,k} = \sum_{\theta,\phi}R_{x,y, c, k}(\theta,\phi)[L_{c}(\theta, \phi) + U_{c}(\theta, \phi)]
\label{Eqn:ceres1}
\vspace{-4pt}
\end{equation}

\begin{figure}[ht]
    \vspace{-5pt}
	\centerline{
		\begin{tabular}{@{}c@{ }c@{}} 
		    \rotatebox{90}{\small{ground truth}} &
			\includegraphics[width=3.1in]{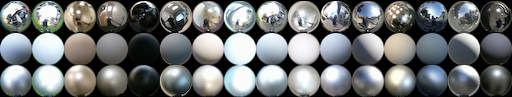} \\
			\rotatebox{90}{\small{renderings}} &
			\includegraphics[width=3.1in]{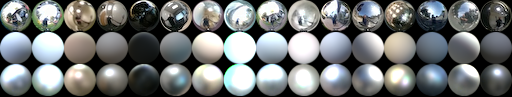} \\
		\end{tabular}}
		\vspace{-10pt}
		\caption{Upper: ground truth LDR sphere images (inputs to the LDR to HDR linear solver). Lower: spheres rendered using the recovered HDR illumination, using image-based relighting and the captured reflectance basis.}
		\vspace{-8pt}
		\label{fig:ceres_solver}
\end{figure}

This generates a set of $m$ linear equations for each BRDF $k$ and color channel $c$, equal to the number of sphere pixels in the reflectance basis images, with $n$ unknown residual light intensities. For lighting directions without clipping, we know that $U_{c}(\theta, \phi)$ = 0. For each color channel, with $km > n$, we can solve for the unknown $U_{c}(\theta, \phi)$ values using non-negative least squares, ensuring light is only added, not removed. In practice, we exclude clipped pixels $p_{x,y,c,k}$ from the solve. Prior methods have recovered clipped light source intensities by comparing the pixel values from a photographed diffuse sphere with the diffuse convolution of a clipped panorama \cite{debevec:2012:single, reinhard:2010:high}, but, to the best of our knowledge, we are the first to use photographed reflectance bases and multiple BRDFs. In Fig. \ref{fig:ceres_solver} upper rows, we show input sphere images extracted from LDR imagery ("ground truth"), and in lower rows, we show the three spheres rendered using Eqn. \ref{Eqn:ceres1}, lit with the HDR illumination recovered from the solver.

We observed when solving for $U_{c}(\theta, \phi)$ treating each color channel independently, brightly-hued red, green, and blue light sources were produced, often at geometrically-nearby lighting directions, rather than a single light source with greater intensity in all three colors channels. To recover results with more plausible, neutrally-colored light sources, we add a cross color channel regularization based on the insight that the color of the photographed diffuse grey ball reveals the average color balance $(R_{avg}, G_{avg}, B_{avg})$ of the bright light sources in the scene. We thus add to our system of equations a new set of linear equations with weight $\lambda=0.5$:

\begin{equation}
\frac{[L_{c=R}(\theta, \phi) + U_{c=R}(\theta, \phi)]}{[L_{c=G}(\theta, \phi) + U_{c=G}(\theta, \phi)]} = 
\frac{R_{avg}}{G_{avg}}
\label{Eqn:ceres2}
\end{equation}

\begin{equation}
\frac{[L_{c=R}(\theta, \phi) + U_{c=R}(\theta, \phi)]}{[L_{c=B}(\theta, \phi) + U_{c=B}(\theta, \phi)]} = 
\frac{R_{avg}}{B_{avg}}
\label{Eqn:ceres3}
\end{equation}

These regularization terms penalize the recovery of strongly-hued light sources of a different color balance than the target diffuse ball. Debevec et al. \shortcite{debevec:2012:single} noted that a regularization term could be added to encourage similar intensities for geometrically-nearby lighting directions, but this would not necessarily prevent the recovery of strongly-hued lights. We recover $U_{c}(\theta, \phi)$ using the Ceres solver \cite{ceres-solver}, promoting our 1 million captured sphere appearances to HDR illumination. As the LDR images from this video-rate data collection method are 8-bit and encoded as sRGB, possibly with local tone-mapping, we first linearize the sphere images assuming $\gamma=2.2$, as required for our linear system formulation. 

\paragraph{Portrait Synthesis.} Using our photographed reflectance fields for each subject and our HDR-promoted lighting, we generate relit portraits with ground truth illumination to serve as training data. We again convert the reflectance basis images into the same relative radiometric space, calibrating based on the incident light source color. As our lighting environments are represented as $32 \times 32$ mirror ball images, we project the reflectance fields onto this basis, again accumulating energy from the input images for each new lighting direction $(\theta, \phi)$ as in LeGendre et al. \shortcite{legendre:2019:deeplight}. Each new basis image is a linear combination of the original 331 OLAT images. 

The lighting capture technique also yields a high-resolution background image corresponding to the three sphere appearances. Since such images on their own contain useful cues for extracting lighting estimates \cite{gardner:2017:indoor, hold:2017:deep}, we composite our relit subjects onto the these backgrounds rather than onto a black frame as in Sun et al. \shortcite{sun:2019:single}, as shown in Fig. \ref{fig:compositing}, producing images which mostly appear to be natural photographs taken out in the wild. Since the background images are 8-bit sRGB, we clip and apply this transfer function to the relit subject images prior to compositing. As in-the-wild portraits are likely to contain clipped pixels (especially for 8-bit live video for mobile AR), we discard HDR data for our relit subjects to match the expected inference-time inputs. 

\begin{figure}[ht]
\vspace{-5pt}
\centering
\includegraphics[width=\linewidth]{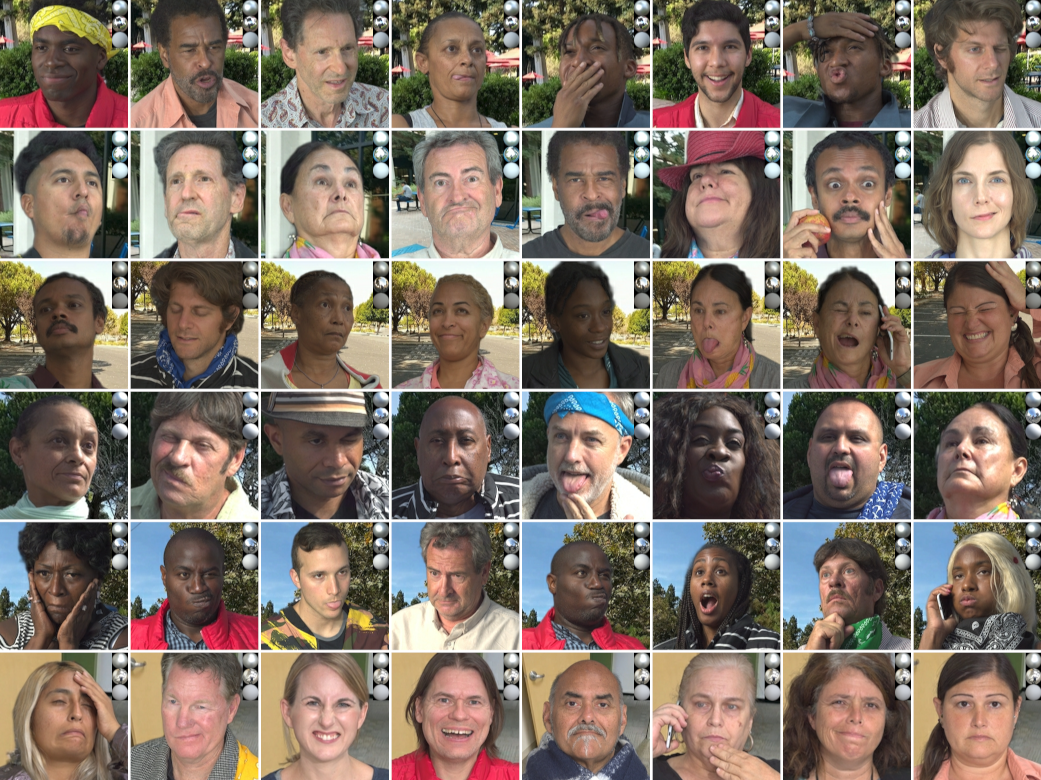}
\vspace{-15pt}
\caption{Example synthetic training portraits, cropped to the bounding box of the detected face. Upper right corners: ground truth HDR illumination for each training example (not included as input during training).}
\label{fig:relit_grid}
\vspace{-15pt}
\end{figure}

\paragraph{Face Localization.} Although background imagery may provide contextual cues that aid in lighting estimation, we do not wish to waste our network's capacity learning a face detector. Instead, we compute a face bounding box for each input, and during training and inference we crop each image, expanding the bounding box by 25\%. During training we add slight crop region variations, randomly changing their position and extent. In our implementation, we use the BlazeFace detector of Bazarevsky et al. \shortcite{bazarevsky:2019:blazeface}, but any could be used. In Fig. \ref{fig:relit_grid} we show example cropped inputs to our model.

\subsection{Network Architecture}
\label{sec:network}
The input to our model is an sRGB encoded LDR image, with the crop of the detected face region of each image resized to an input resolution of $256 \times 256$ and normalized to the range of $[-0.5, 0.5]$. We use an encoder-decoder architecture with a latent vector of size 1024 at the bottleneck, representing log-space HDR illumination, as the sun can be several orders of magnitude brighter than the sky \cite{stumpfel:2004:direct}. The encoder consists of five $3 \times 3$ convolutions each followed by a blur-pooling operation \cite{zhang:2019:making}, with successive filter depths of 16, 32, 64, 128, and 256, followed by one last convolution with a filter size of $8 \times 8$ and depth 256, and finally a fully-connected layer. The decoder consists of three sets of $3 \times 3$ convolutions of filter depths 64, 32, and 16, each followed by a bilinear-upsampling operation. The final output of the network is a $32 \times 32$ HDR image of a mirror ball representing log-space omnidirectional illumination.

We also use an auxiliary discriminator architecture to add an adversarial loss term, enforcing estimation of plausible high frequency illumination (see Sec. \ref{sec:loss_function}). This network takes as input clipped images of ground truth and predicted illumination from the main model, and tries to discriminate between the real and generated examples. The discriminator encoder consists of three $3 \times 3$ convolutions each followed by a max-pooling operation, with successive filter depths of 64, 128, and 256, followed by a fully connected layer of size 1024 before the final output layer. As our main network's decoder includes several upsampling operations, our network is implicitly learning information at multiple scales. We leverage this multi-scale output to provide inputs to the discriminator not just of the full-resolution $32 \times 32$ clipped lighting image, but also of a lighting image at each scale: $4 \times 4$, $8 \times 8$, and $16 \times 16$, using the multi-scale gradient technique of MSG-GAN \cite{karnewar:2020:msg}. As the lower-resolution feature maps produced by our generator network have more than 3 channels, we add a convolution operation at each scale as extra branches of the network, producing multiple scales of 3-channel lighting images to supply to the discriminator.

\subsection{Loss Function}
\label{sec:loss_function}

\paragraph{Multi-scale Image-Based Relighting Rendering Loss.} LeGendre et al. \shortcite{legendre:2019:deeplight} describe a differentiable image-based relighting rendering loss, used for training a network to estimate HDR lighting $\hat{L}$ from an unconstrained image. This approach minimizes the reconstruction loss between the ground truth sphere images $I$ for multiple BRDFs and the corresponding network-rendered spheres $\hat{I}$, lit with the predicted illumination. We use this technique to train our model for inverse lighting from portraits, relying on these sphere renderings to learn illumination useful for rendering virtual objects of a variety of BRDFs. We produce sphere renderings $\hat{I}$ in-network using image-based relighting and photographed reflectance fields for each sphere of BRDF index $k$ (mirror, matte silver, or diffuse), and color channel $c$, with $\hat{L_{c}}(\theta,\phi)$ as the intensity of light for the direction $(\theta,\phi)$:
\vspace{-3pt}
\begin{equation}
\hat{I}_{x,y,k,c} = \sum_{\theta,\phi}R_{x,y,k,c}(\theta,\phi) \hat{L_{c}}(\theta,\phi).
\label{Eqn:IBRL}
\end{equation}

As in LeGendre et al. \shortcite{legendre:2019:deeplight}, our network similarly outputs a log space image $Q$ of HDR illumination, with pixel values $Q_{c}(\theta, \phi)$, so sphere images are rendered as:
\vspace{-3pt}
\begin{equation}
\hat{I}_{x,y,k,c} = \sum_{\theta,\phi}R_{x,y,k,c}(\theta,\phi) e^{Q_{c}(\theta,\phi)}.
\label{Eqn:LogIBRL}
\end{equation}

With binary mask $\hat M$ to mask out the corners of each sphere, $\gamma=2.2$ for gamma-encoding, $\lambda_k$ as an optional weight for each BRDF, and a differentiable soft-clipping function $\Lambda$ as in LeGendre et al. \shortcite{legendre:2019:deeplight}, the final LDR image reconstruction loss $L_\text{rec}$ comparing ground truth images $I_k$ and network-rendered images $\hat{I_k}$ is:
\vspace{-3pt}
\begin{equation}
L_\text{rec} = \sum_{k=0}^{2}\lambda_{k}
\big\|\hat M\odot (\Lambda({\hat I_{k}})^{\tfrac{1}{\gamma}} - \Lambda(I_{k}))\big\|_{1}.
\label{Eqn:L1}
\end{equation}

Rather than use the LDR sphere images captured in the video-rate data collection as the reference images $I_{k}$, we instead render the spheres with the HDR lighting recovered from the linear solver of Sec. \ref{sec:data}, gamma-encoding the renderings with $\gamma=2.2$. This ensures that the same lighting is used to render the "ground truth" spheres as the input portraits, preventing the propagation of residual error from the HDR lighting recovery to our model training phase.

We finally add extra convolution branches to convert the multi-scale feature maps of the decoder into 3-channel images representing log-space HDR lighting at successive scales. We then extend the rendering loss function of LeGendre et al. \shortcite{legendre:2019:deeplight} (Eqn. \ref{Eqn:L1}) to the multi-scale domain, rendering mirror, matte silver, and diffuse spheres during training in sizes $4 \times 4$, $8 \times 8$, $16 \times 16$, and $32 \times 32$. With scale index represented by $s$, and an optional weight for each as $\lambda_s$, our multi-scale image reconstruction loss is written as:
\vspace{-5pt}
\begin{equation}
L_\text{ms-rec} = \sum_{s=0}^{3}\sum_{k=0}^{2}\lambda_{s}\lambda_{k}
\big\|\hat M\odot (\Lambda({\hat I_{k}})^{\tfrac{1}{\gamma}} - \Lambda(I_{k}))\big\|_{1}.
\label{Eqn:L1_multiscale}
\vspace{-2pt}
\end{equation}

\paragraph{Adversarial Loss.} Recent work in unconstrained lighting estimation has shown that adversarial loss terms improve the recovery of high-frequency information compared with using only image reconstruction losses \cite{legendre:2019:deeplight, song:2019:neural}. Thus, we add an adversarial loss term with weight $\lambda_{adv}$ as in LeGendre et al. \shortcite{legendre:2019:deeplight}. However, in contrast to this technique, we use a multi-scale GAN architecture that flows gradients from the discriminator to the generator network at multiple scales \cite{karnewar:2020:msg}, providing the discriminator with different sizes of both real and generated clipped mirror ball images.

\subsection{Implementation Details}
\label{sec:implementation}

We use Tensorflow and the ADAM \cite{kinga:2015:adam} optimizer with $\beta_{1}=0.9$, $\beta_{2}=0.999$, a learning rate of $0.00015$ for the generator network, and, as is common, one 100$\times$ lower for the discriminator network, alternating between training the generator and discriminator. We set $\lambda_{k}=0.2, 0.6, 0.2$ for the mirror, diffuse, and matte silver BRDFs respectively, set $\lambda_{s}=1$ to weight all image scales equally, set $\lambda_{adv}=0.004$, and use a batch size of 32. As the number of lighting environments is orders of magnitude larger than the number of subjects, we found that early stopping at $1.2$ epochs appears to prevent over-fitting to subjects in the training set. We use the ReLU activation function for the generator network and the ELU activation function \cite{clevert:2016:elu} for the discriminator. To augment our dataset, we flip both the input images and lighting environments across the vertical axis. 

\paragraph{Datasets.} We split our 70 subjects into two groups: 63 for training and 7 for evaluation, ensuring that all expressions and camera views for a given subject belong to the same subset. We manually select the 7 subjects to include both skin tone and gender diversity. In total, for each of our 1 million lighting environments, we randomly select 8 OLAT sequences to relight from the training set (across subjects, facial expressions, and camera views), generating a training dataset of 8 million portraits with ground truth illumination (examples in Fig. \ref{fig:relit_grid}). Using the same method, we capture lighting environments in both indoor and outdoor locations unseen in training to use for our evaluation, pairing these only with the evaluation subjects.

\section{Evaluation}

In this section, we compare against prior techniques, perform an ablation study to investigate the performance gains for various sub-components, and measure performance across our diverse evaluation subjects. We also use our lighting estimates to render and composite virtual objects into real-world imagery.

\subsection{Comparisons}
\label{sec:comparisons}

\begin{figure*}[t]
\vspace{-5pt}
\centering
\includegraphics[width=\linewidth]{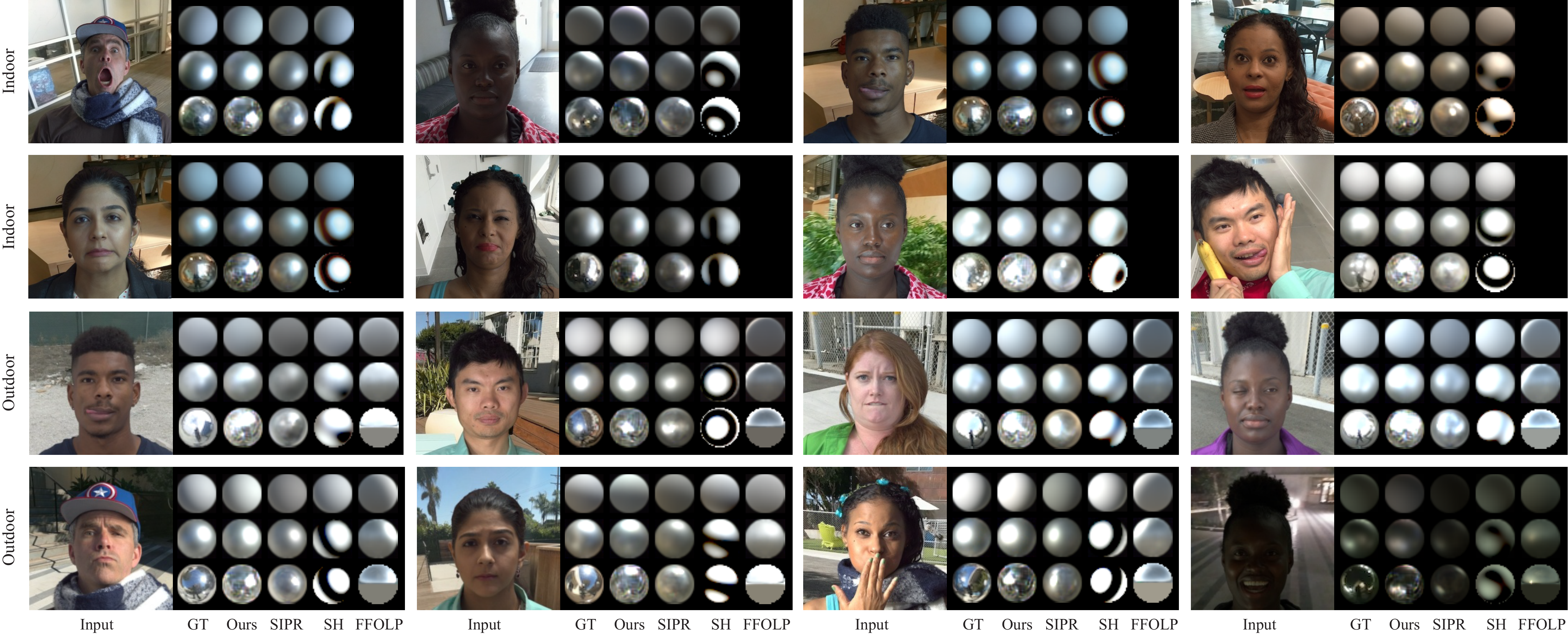}
\vspace{-20pt}
\caption{Comparison sphere renderings (diffuse, matte silver, and mirror) for evaluation subjects and indoor and outdoor lighting environments. We compare our method against "Single Image Portrait Relighting" (SIPR) \cite{sun:2019:single}, the second order SH decomposition of the ground truth illumination, and for outdoor scenes, a radiance-scaled version of "From Faces to Outdoor Light Probes" (FFOLP) \cite{calian:2018:faces}. Our model more faithfully recovers the total scene radiance compared with SIPR, and, unlike the SH decomposition, is useful for rendering materials with BRDFs beyond Lambertian.}
\label{fig:comparison_grid}
\vspace{-5pt}
\end{figure*}

Accurately estimated lighting should correctly render objects with arbitrary reflectance properties, so we test our model's performance using $L_\text{rec}$. This metric compares the appearance of three spheres (diffuse, matte silver, and mirror) as rendered with the ground truth versus estimated illumination. In Table \ref{Table:comparison_L1}, we compare our model against Sun et al. \shortcite{sun:2019:single}, Calian et al. \shortcite{calian:2018:faces}, and a 2\textsuperscript{nd} order SH decomposition of the ground truth lighting. We use our own implementation for Sun et al. \shortcite{sun:2019:single}, training the model on our data for a fair comparison. As in the original implementation, we train the model with random crops from portraits composited over black backgrounds (not real-world imagery). As the method includes loss terms on both relit portraits and lighting, we generate 4 million portrait pairs from our original images and train the joint portrait relighting / lighting estimation model. To compare with Calian et al. \shortcite{calian:2018:faces}, the authors generously computed outdoor lighting for a set of portraits. However, the scale of their lighting depends on an albedo prior fit to a different dataset. So, for a best case comparison, we re-scale the author-provided illumination such that the total scene radiance matches that of the ground truth. Finally, we compare against the 2\textsuperscript{nd} order SH decomposition, as this represents the best case scenario for any monocular face reconstruction technique that models illumination with this low frequency basis. 

For the LDR image reconstruction losses, our model out-performs Sun et al. \shortcite{sun:2019:single} and Calian et al. \shortcite{calian:2018:faces} for the diffuse and matte silver spheres. However, Sun et al. \shortcite{sun:2019:single} out-performs ours for the mirror sphere, as it its log-space loss on lighting is similar to $L_\text{rec}$ for the mirror ball (but in HDR). As expected, the 2\textsuperscript{nd} order SH approximation of the ground truth illumination out-performs our model for $L_\text{rec}$ for the diffuse ball, since a low frequency representation of illumination suffices for rendering Lambertian materials. However, our model out-performs the 2\textsuperscript{nd} order SH decomposition for $L_\text{rec}$ for both the matte silver and mirror balls, with non-Lambertian BRDFs. This suggests that lighting produced by our model is better suited for rendering diverse materials.

\begin{table}[h]
\vspace{-5pt}
\caption{\small Comparison among methods: Average $L_{1}$ loss by BRDF [diffuse (d), mirror (m), and matte silver (s) spheres (in columns)], for evaluation portraits. We compare ground truth sphere images with those \textit{rendered} using the HDR lighting inference, for \textit{unseen} indoor (UI) and outdoor (UO) locations. (*$n=237$ for Calian et al. \shortcite{calian:2018:faces} due to face tracking failures.)}
\vspace{-5pt}
\footnotesize
\centering
\begin{tabular}{@{}l@{\quad}c@{\quad}c@{}c@{\quad\;}c@{\quad}c@{}c@{\quad\;}c@{\quad}c@{}c@{}}
\toprule
 & \multicolumn{2}{@{}c@{}}{$L_{1(d)}$} & & \multicolumn{2}{@{}c@{}}{$L_{1(s)}$} & & \multicolumn{2}{@{}c@{}}{$L_{1(m)}$}  \\
\cmidrule{2-3} \cmidrule{5-6} \cmidrule{8-9}
$n=270$\textsuperscript{*} & UI & UO & & UI & UO & & UI & UO  \\
\midrule
Our model  
        & \textbf{0.069} & \textbf{0.056} & & \textbf{0.087} & \textbf{0.072} & & 0.181 & 0.157 \\
2\textsuperscript{nd} order SH of GT 
        & \textbf{0.016} & \textbf{0.015} & & 0.120 & 0.109 & & 0.306 & 0.247 \\
Sun et al. \shortcite{sun:2019:single}       
        & 0.145 & 0.120 & & 0.113 & 0.100 & & \textbf{0.154} & \textbf{0.139} \\
Calian et al. \shortcite{calian:2018:faces}
        & -- & 0.158 & & -- & 0.163 & & -- & 0.215 \\
\bottomrule
\end{tabular}
\vspace{-5pt}
\label{Table:comparison_L1}
\end{table}

In Table \ref{Table:comparison_radiance}, we compare the relative radiance for each color channel for our model and that of Sun et al. \shortcite{sun:2019:single}, computed as the sum of the pixels of the predicted illumination subtracted from the ground truth illumination, divided by the sum of the ground truth. We show that on average, the illumination recovered by the method of Sun et al. \shortcite{sun:2019:single} is missing 41\% of the scene radiance. In contrast, for this randomly selected evaluation subset, our method adds on average 9\% to the total scene radiance. As our rendering-based loss terms include matching the appearance of a diffuse ball, which is similar to a diffuse convolution of the HDR lighting environment, our method is able to more faithfully recover the total scene radiance. 

\begin{table}[h]
\caption{\small Average relative radiance difference [(GT - Pred) / GT] for estimated lighting, comparing our method and Sun et al. \shortcite{sun:2019:single}.}
\vspace{-5pt}
\footnotesize
\centering
\begin{tabular}{@{}l@{\quad}c@{\quad}c@{}c@{\quad\;}c@{\quad}c@{}c@{\quad\;}c@{\quad}c@{}c@{}}
\toprule
 & \multicolumn{2}{@{}c@{}}{Red Channel} & & \multicolumn{2}{@{}c@{}}{Green Channel} & & \multicolumn{2}{@{}c@{}}{Blue Channel}  \\
\cmidrule{2-3} \cmidrule{5-6} \cmidrule{8-9}
$n=270 $  & UI & UO & & UI & UO & & UI & UO  \\
\midrule
Our model  
        & \textbf{-9.04\%} & \textbf{-6.22\%} & & \textbf{-6.22\%} & \textbf{-6.10\%} & & \textbf{-7.66\%} & \textbf{-17.88\%} \\
Sun et al. \shortcite{sun:2019:single}       
        & 34.53\% & 41.79\% & & 38.31\% & 44.55\% & & 39.73\% & 48.19\% \\
\bottomrule
\end{tabular}
\vspace{-8pt}
\label{Table:comparison_radiance}
\end{table}

In Fig. \ref{fig:comparison_grid} we show qualitative results, rendering the three spheres using illumination produced using our method, that of Sun et al. \shortcite{sun:2019:single} labeled as "SIPR", that of a 2\textsuperscript{nd} order SH decomposition, and that of Calian et al. \shortcite{calian:2018:faces} for outdoor scenes, labeled as "FFOLP". The missing scene radiance from the method of Sun et al. \shortcite{sun:2019:single} is apparent looking at the diffuse sphere renderings, which are considerably darker than ground truth for this method. While the 2\textsuperscript{nd} order SH approximation of the ground truth lighting produces diffuse sphere renderings nearly identical to the ground truth, Fig. \ref{fig:comparison_grid} again shows how this approximation is ill-suited to rendering non-Lambertian materials. For the method of Calian et al. \shortcite{calian:2018:faces}, the sun direction is misrepresented as our evaluation lighting environments include a diversity of camera elevations, with the horizon line not exclusively along the equator of the mirror sphere. 

\begin{figure*}[t]
    \scriptsize
    \vspace{-6pt}
	\centerline{
		\begin{tabular}{c@{ }c@{ }c@{ }c@{ }c@{ }c@{ }c@{}}
			\includegraphics[height=1.05in]{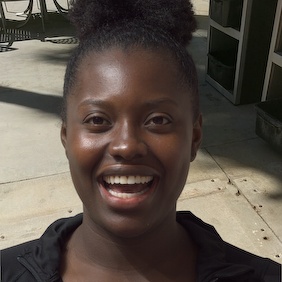}&
			\includegraphics[height=1.05in]{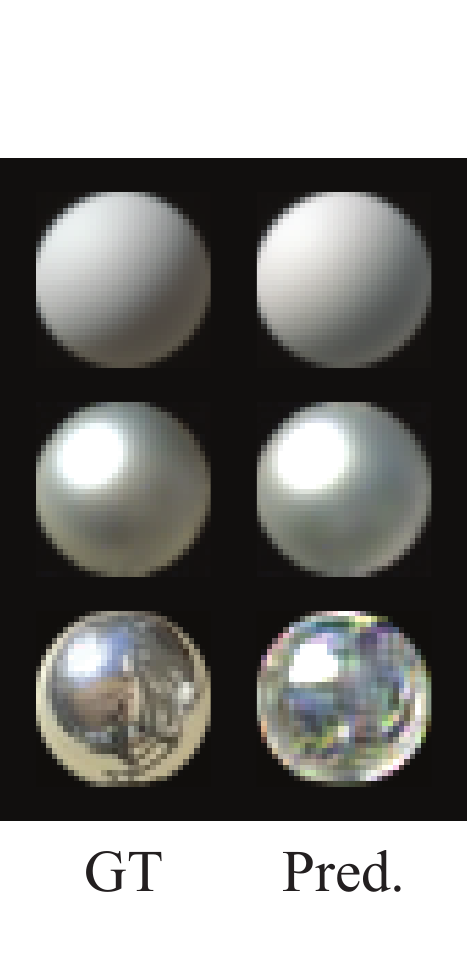} &
			\includegraphics[height=1.05in]{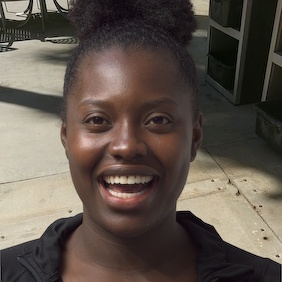} &
			\includegraphics[height=1.05in]{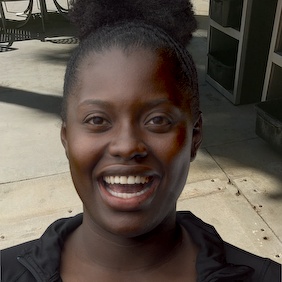} &
			\includegraphics[height=1.05in]{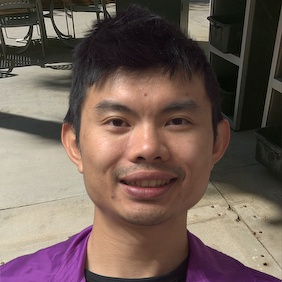} &
			\includegraphics[height=1.05in]{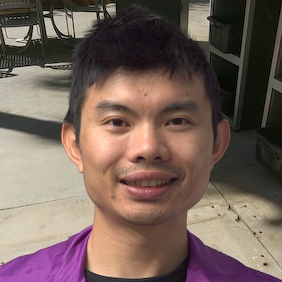} &
			\includegraphics[height=1.05in]{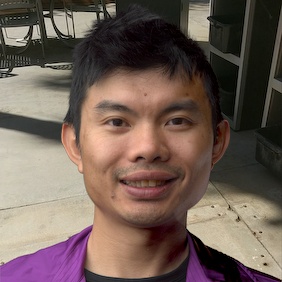}  \\
			\includegraphics[height=1.05in]{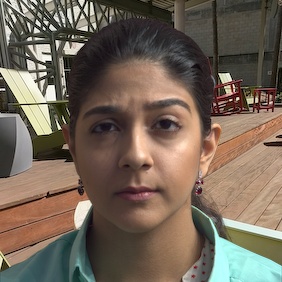}&
			\includegraphics[height=1.05in]{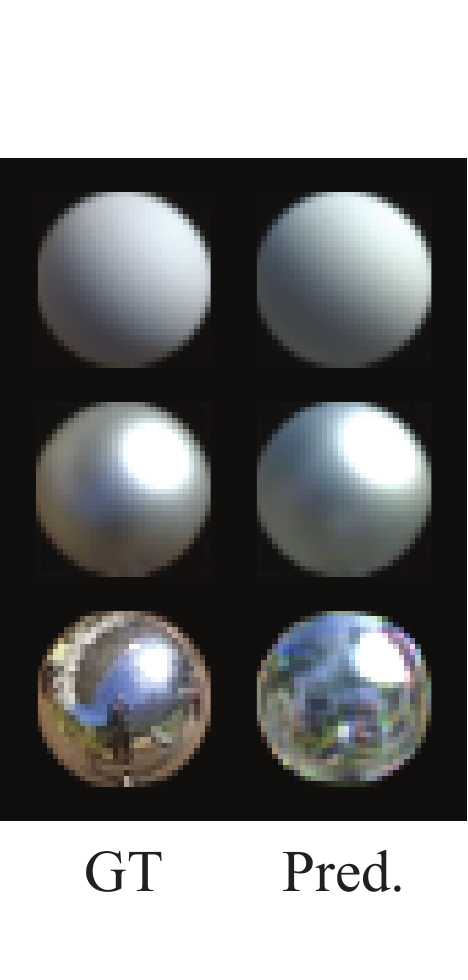} &
			\includegraphics[height=1.05in]{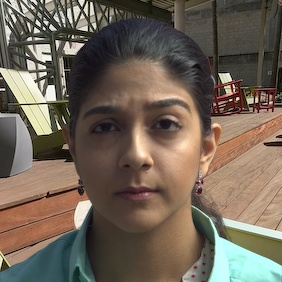} &
			\includegraphics[height=1.05in]{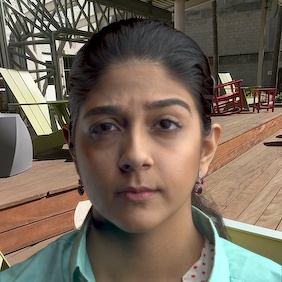} &
			\includegraphics[height=1.05in]{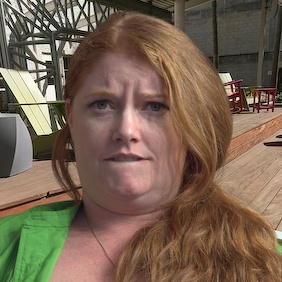} &
			\includegraphics[height=1.05in]{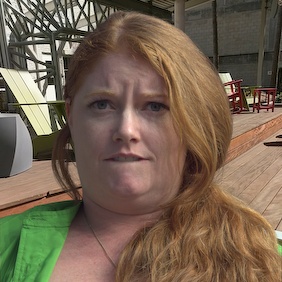} &
			\includegraphics[height=1.05in]{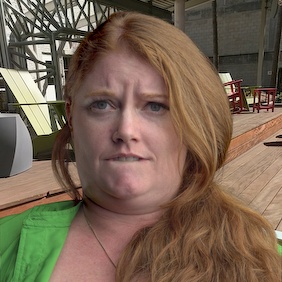}  \\
			\includegraphics[height=1.05in]{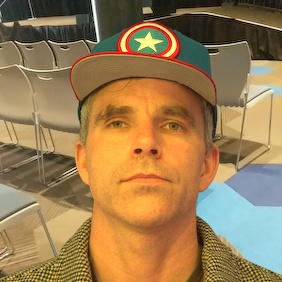}&
			\includegraphics[height=1.05in]{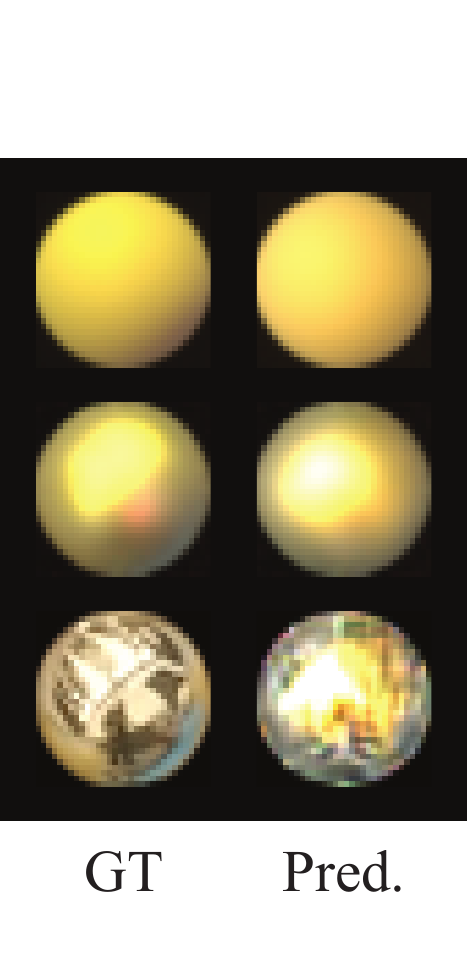} &
			\includegraphics[height=1.05in]{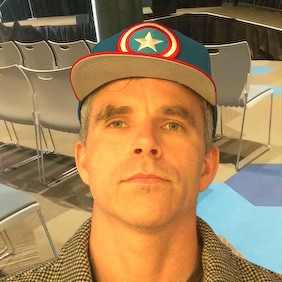} &
			\includegraphics[height=1.05in]{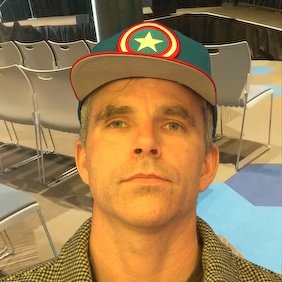} &
			\includegraphics[height=1.05in]{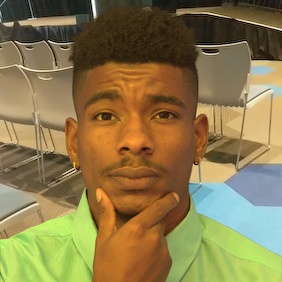} &
			\includegraphics[height=1.05in]{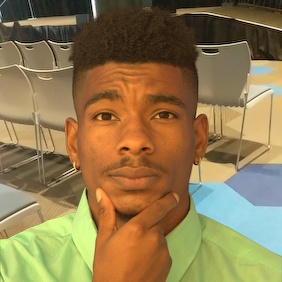} &
			\includegraphics[height=1.05in]{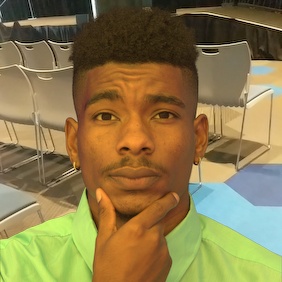}  \\
			\includegraphics[height=1.05in]{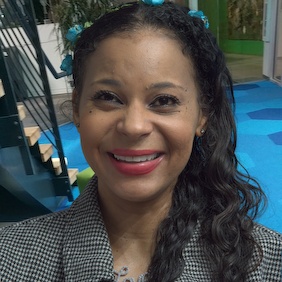}& 
			\includegraphics[height=1.05in]{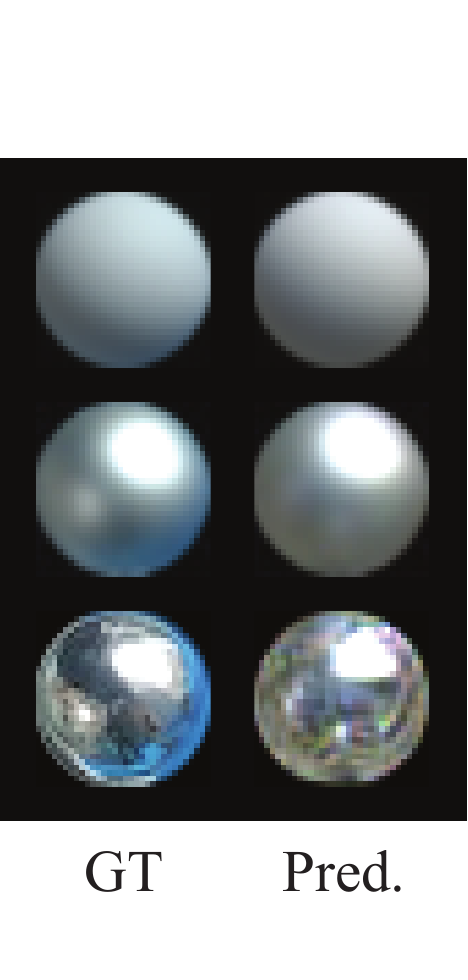} & 
			\includegraphics[height=1.05in]{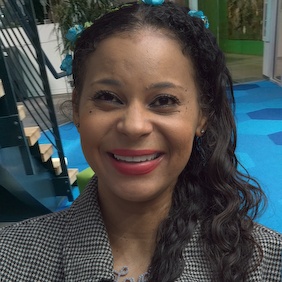} & 
			\includegraphics[height=1.05in]{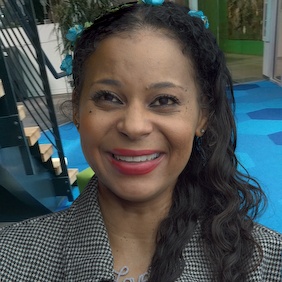} &
			\includegraphics[height=1.05in]{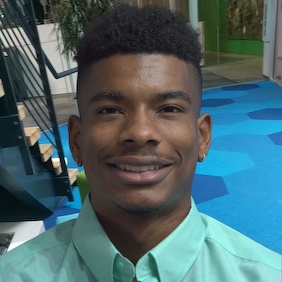} & 
			\includegraphics[height=1.05in]{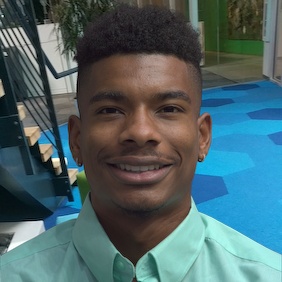} & 
			\includegraphics[height=1.05in]{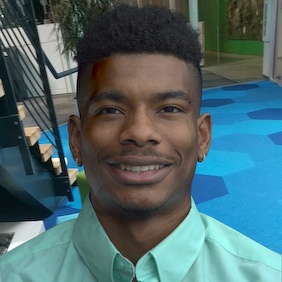}  \\ 
			\footnotesize{\textbf{(a)} input image} & 
			\footnotesize{\textbf{(b)} lighting} &
			\footnotesize{\textbf{(c)} lit with pred. from \textbf{a}} &
			\footnotesize{\textbf{(d)} lit with SH of GT} &
			\footnotesize{\textbf{(e)} lit with GT} &
			\footnotesize{\textbf{(f)} lit with pred. from \textbf{a}} &
			\footnotesize{\textbf{(g)} lit with SH of GT} \\
		\end{tabular}}
		\vspace{-5pt}
		\caption{\textbf{(a)} Inputs to our model, generated using image-based relighting and a photographed reflectance basis for each evaluation subject. \textbf{(b)} Left: ground truth (GT) lighting used to generate \textbf{a}; Right: lighting estimated from \textbf{a} using our method. \textbf{(c)} The same subject lit with the predicted lighting. \textbf{(d)} The same subject lit with the 2\textsuperscript{nd} order SH decomposition of the GT lighting. \textbf{(e)} A new subject lit with the GT lighting. \textbf{(f)} The new subject lit with the illumination estimated from \textbf{a} using our method. \textbf{(g)} The new subject lit with the 2\textsuperscript{nd} order SH decomposition of the GT lighting. Our method produces lighting environments that can be used to realistically render virtual subjects into existing scenes, while the 2\textsuperscript{nd} order SH lighting leads to an overly diffuse skin appearance.}
		\vspace{-5pt}
		\label{fig:relighting_grid}
\end{figure*}

In Fig. \ref{fig:relighting_grid}, we show an example where the illumination is estimated from a synthetic LDR portrait of a given subject (Fig. \ref{fig:relighting_grid}\textbf{a}), with the estimated and ground truth illumination in Fig. \ref{fig:relighting_grid}\textbf{b}. We then use both the estimated illumination from our model and the 2\textsuperscript{nd} order SH approximation of the ground truth to light the same subject, shown in Fig. \ref{fig:relighting_grid}\textbf{c} and \textbf{d} respectively. For lighting environments with high frequency information (rows 1, 2, and 4 in Fig. \ref{fig:relighting_grid}), our lighting estimates produce portraits that more faithfully match the input images. These results highlight the limitation inherent in the Lambertian skin reflectance assumption. 

\subsection{Ablation Study}
\label{sec:ablation}

\begin{figure}[ht]
\vspace{-3pt}
\centering
\includegraphics[width=\linewidth]{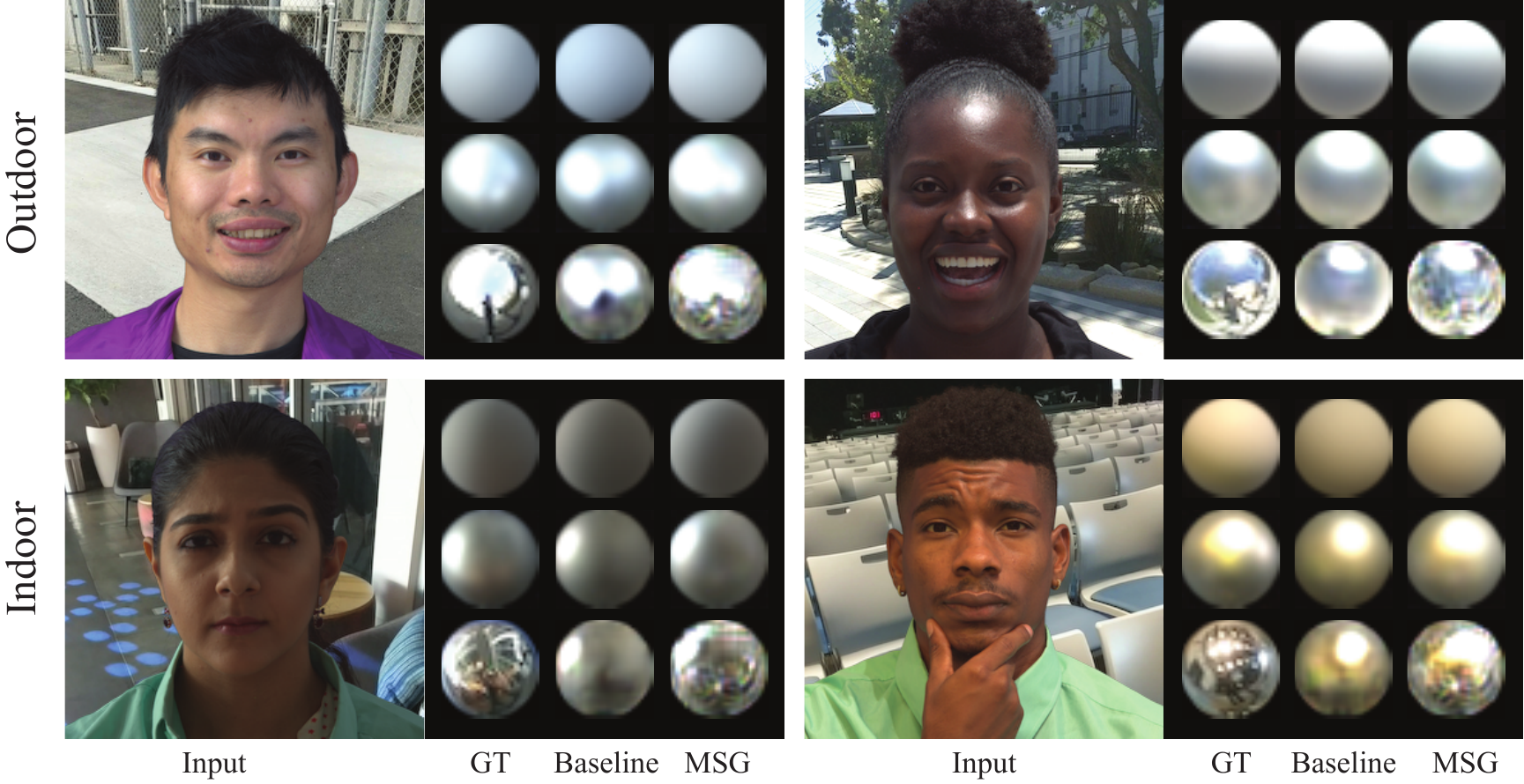}
\vspace{-15pt}
\caption{Our multi-scale losses increase the sharpness of features in the recovered illumination, as shown in the mirror ball images (bottom rows), compared with baseline. Upper-right grid shown at +1 stop for display.}
\vspace{-8pt}
\label{fig:ablation_grid_image}
\end{figure}

In Table \ref{Table:ablation}, we report $L_\text{rec}$ for each BRDF when evaluating each component of our system. We compare a baseline model using the single-scale losses \cite{legendre:2019:deeplight} to our proposed model trained with multi-scale losses ($L_\text{ms-rec}$ and MSG-GAN). The multi-scale loss modestly decreases $L_\text{rec}$ for both the diffuse and matte silver spheres, while increasing that of the mirror sphere. This increase is expected, as the adversarial loss for the mirror ball pulls the estimate away from an overly-blurred image that minimizes $L_\text{rec}$. In Fig. \ref{fig:ablation_grid_image}, we show the visual impact of the multi-scale loss term, which synthesizes more high frequency details. 

\begin{table}[h]
\vspace{-3pt}
\caption{\small Average $L_{1}$ loss by BRDF: diffuse (d), mirror (m), and matte silver (s) spheres (in columns), for lighting estimated from portraits of our evaluation subjects, using our technique with and without different features.}
\vspace{-5pt}
\footnotesize
\centering
\begin{tabular}{@{}l@{\quad}c@{\quad}c@{}c@{\quad\;}c@{\quad}c@{}c@{\quad\;}c@{\quad}c@{}c@{}}
\toprule
 & \multicolumn{2}{@{}c@{}}{$L_{1(d)}$} & & \multicolumn{2}{@{}c@{}}{$L_{1(s)}$} & & \multicolumn{2}{@{}c@{}}{$L_{1(m)}$}  \\
\cmidrule{2-3} \cmidrule{5-6} \cmidrule{8-9}
Model, $n=3968$  & UI & UO & & UI & UO & & UI & UO  \\
\midrule
Proposed (no Multi-scale Losses)                      & 0.054 & 0.050 & & 0.076 & 0.069 & & \textbf{0.144} & \textbf{0.128} \\
No Face Detector            
                              & 0.055 & 0.051 & & 0.080 & 0.075 & & 0.151 & 0.136 \\
No Background Imagery        
                              & 0.057 & 0.053 & & 0.078 & 0.072 & & 0.147 & 0.133 \\
Proposed with Multi-scale Losses
                              & \textbf{0.050} & \textbf{0.047} & & \textbf{0.072} & \textbf{0.067} & & 0.156 & 0.141 \\
log-L\textsubscript{2} Loss (as in Sun et al. \shortcite{sun:2019:single}) 
                              & 0.151 & 0.133 & & 0.114 & 0.103 & & 0.152 & 0.132 \\
No Face (LeGendre et al. \shortcite{legendre:2019:deeplight})     
                              & 0.136 & 0.135 & & 0.144 & 0.137 & & 0.174 & 0.166 \\
\bottomrule
\end{tabular}
\vspace{-8pt}
\label{Table:ablation}
\end{table} 

In Table \ref{Table:ablation}, we also compare our baseline model, trained on images cropped using a face detector, to a model trained on random crops as in Sun et al. \shortcite{sun:2019:single}, labeled "No Face Detector." The face detector imparts some modest improvement. Additionally, we compare our baseline model, trained on portraits composited onto real-world background imagery matching the ground truth illumination, to one trained without backgrounds, with subjects composited instead over black as in Sun et al. \shortcite{sun:2019:single}. (The evaluation in this case is also performed on subjects against black backgrounds). The backgrounds also impart some modest improvement. We further show that our baseline model outperforms a model trained using the log-L\textsubscript{2} loss on HDR lighting of Sun et al. \shortcite{sun:2019:single}. As this loss function does not include a rendering step, this is somewhat expected. Finally, we compare against a model trained using \textit{only} random crops of the background imagery, without portraits, using the single-scale loss terms. This table entry, labeled as "No Face," is equivalent to LeGendre et al. \shortcite{legendre:2019:deeplight}, but trained on our background images and with our network architecture. As expected, the presence of faces in the input images significantly improves model performance.

\subsection{Lighting Consistency for Diverse Skin Tones}
\label{sec:skin_tones}

\begin{figure}[ht]
\vspace{-6pt}
\centering
\includegraphics[width=\linewidth]{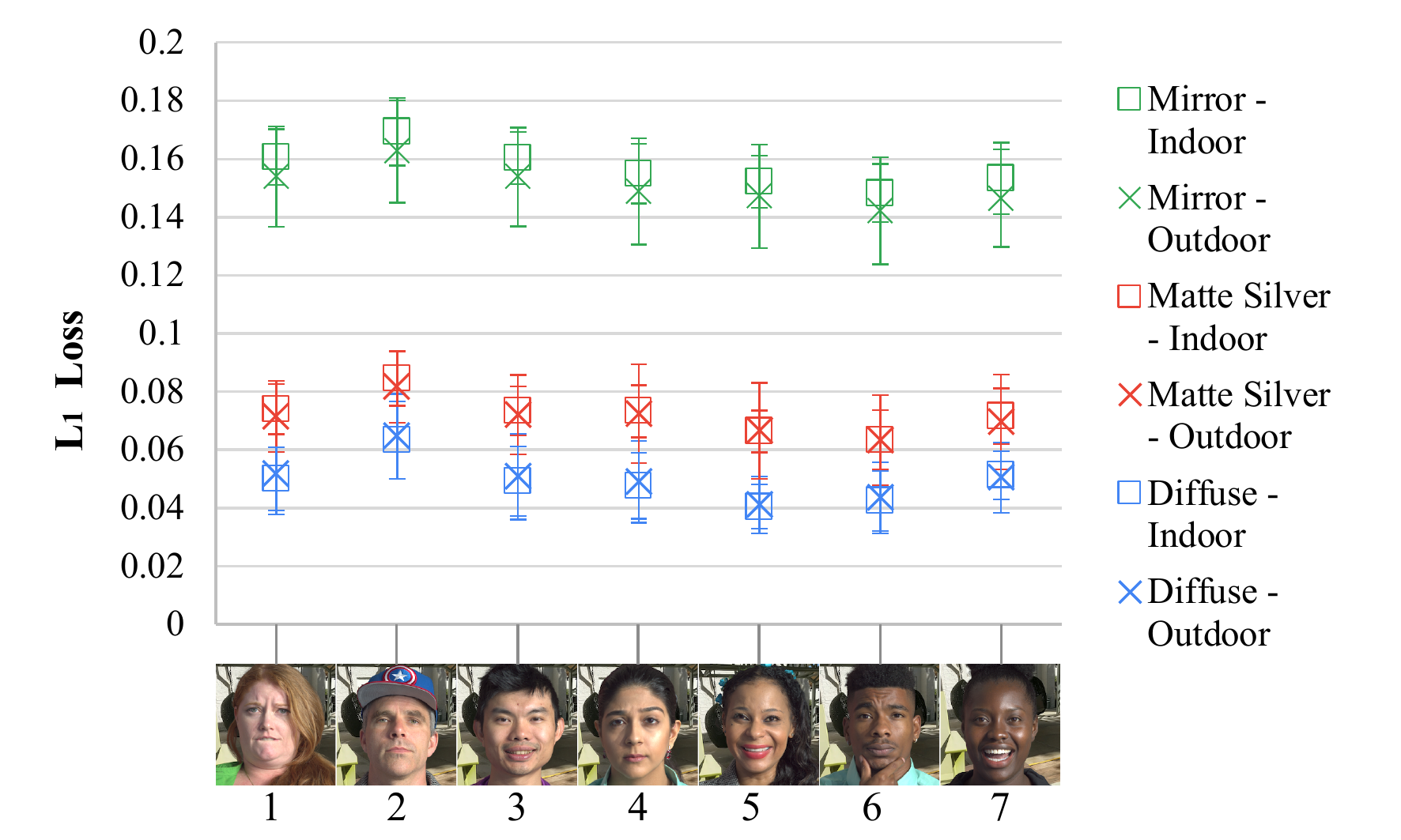}
\vspace{-17pt}
\caption{Average $L_\text{rec}$ for individual evaluation subjects, with $n=496$ each for unseen indoor and outdoor scenes. This plot corresponds with Table. \ref{Table:ml_fairness}. Our model's performance is similar for subjects of diverse skin tones.}
\label{fig:ml_fairness_plot}
\vspace{-10pt}
\end{figure}

In Table \ref{Table:ml_fairness}, we report $L_\text{rec}$ for each of the three spheres individually, for $496$ test examples in unseen indoor and outdoor lighting environments for each evaluation subject. Each example set includes diverse camera viewpoints, facial expressions, and hats/accessories. In Fig. \ref{fig:ml_fairness_plot}, we plot the data of Table \ref{Table:ml_fairness} to visualize that while there are some slight variations in $L_\text{rec}$ across subjects, the model's performance appears similar across diverse skin tones. 

\begin{table}[h]
\caption{\small Average $L_{1}$ loss by BRDF: diffuse (d), mirror (m), and matte silver (s) spheres (in columns), for lighting estimated from portraits of our evaluation subjects, numbered 1-7 (see Fig. \ref{fig:ml_fairness_plot}). This table corresponds with Fig. \ref{fig:ml_fairness_plot}.}
\vspace{-5pt}
\footnotesize
\centering
\begin{tabular}{@{}l@{\quad}c@{\quad}c@{}c@{\quad\;}c@{\quad}c@{}c@{\quad\;}c@{\quad}c@{}c@{}}
\toprule
 & \multicolumn{2}{@{}c@{}}{$L_{1(d)}$} & & \multicolumn{2}{@{}c@{}}{$L_{1(s)}$} & & \multicolumn{2}{@{}c@{}}{$L_{1(m)}$}  \\
\cmidrule{2-3} \cmidrule{5-6} \cmidrule{8-9}
$n=496$  & UI & UO & & UI & UO & & UI & UO  \\
\midrule
Subject 1  & 0.050 & 0.052 & & 0.074 & 0.071 & & 0.161 & 0.154 \\
Subject 2  & 0.063 & 0.065 & & 0.084 & 0.081 & & 0.169 & 0.162 \\
Subject 3  & 0.049 & 0.051 & & 0.073 & 0.072 & & 0.160 & 0.154 \\
Subject 4  & 0.048 & 0.049 & & 0.073 & 0.072 & & 0.155 & 0.149 \\
Subject 5  & 0.040 & 0.041 & & 0.066 & 0.066 & & 0.152 & 0.147 \\
Subject 6  & 0.042 & 0.043 & & 0.063 & 0.063 & & 0.148 & 0.142 \\
Subject 7  & 0.051 & 0.050 & & 0.071 & 0.070 & & 0.153 & 0.146 \\
\bottomrule
\end{tabular}
\vspace{-7pt}
\label{Table:ml_fairness}
\end{table}

\begin{figure}[ht]
\vspace{-12pt}
\centering
\includegraphics[width=\linewidth]{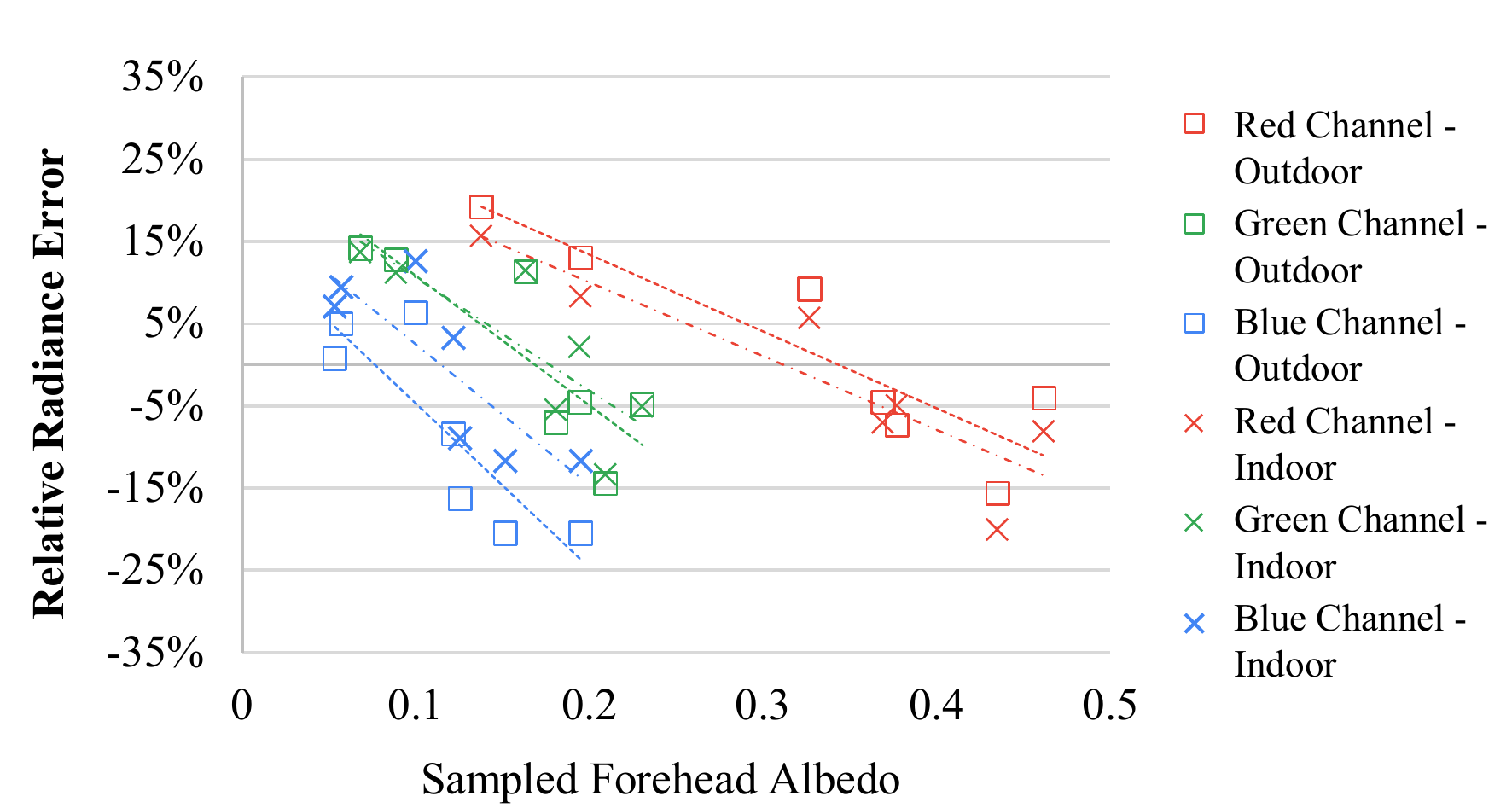}
\vspace{-16pt}
\caption{y axis: Average relative error in total radiance [(GT-Pred.)/GT] for each color channel for each of our evaluation subjects ($n=496$ each for unseen indoor and outdoor scenes). x axis: Each subject's average RGB albedo, sampled from the forehead under a unit sphere of illumination.}
\label{fig:relative_radiance}
\vspace{-10pt}
\end{figure}

\begin{figure}[ht]
\centering
\vspace{-3pt}
\includegraphics[width=\linewidth]{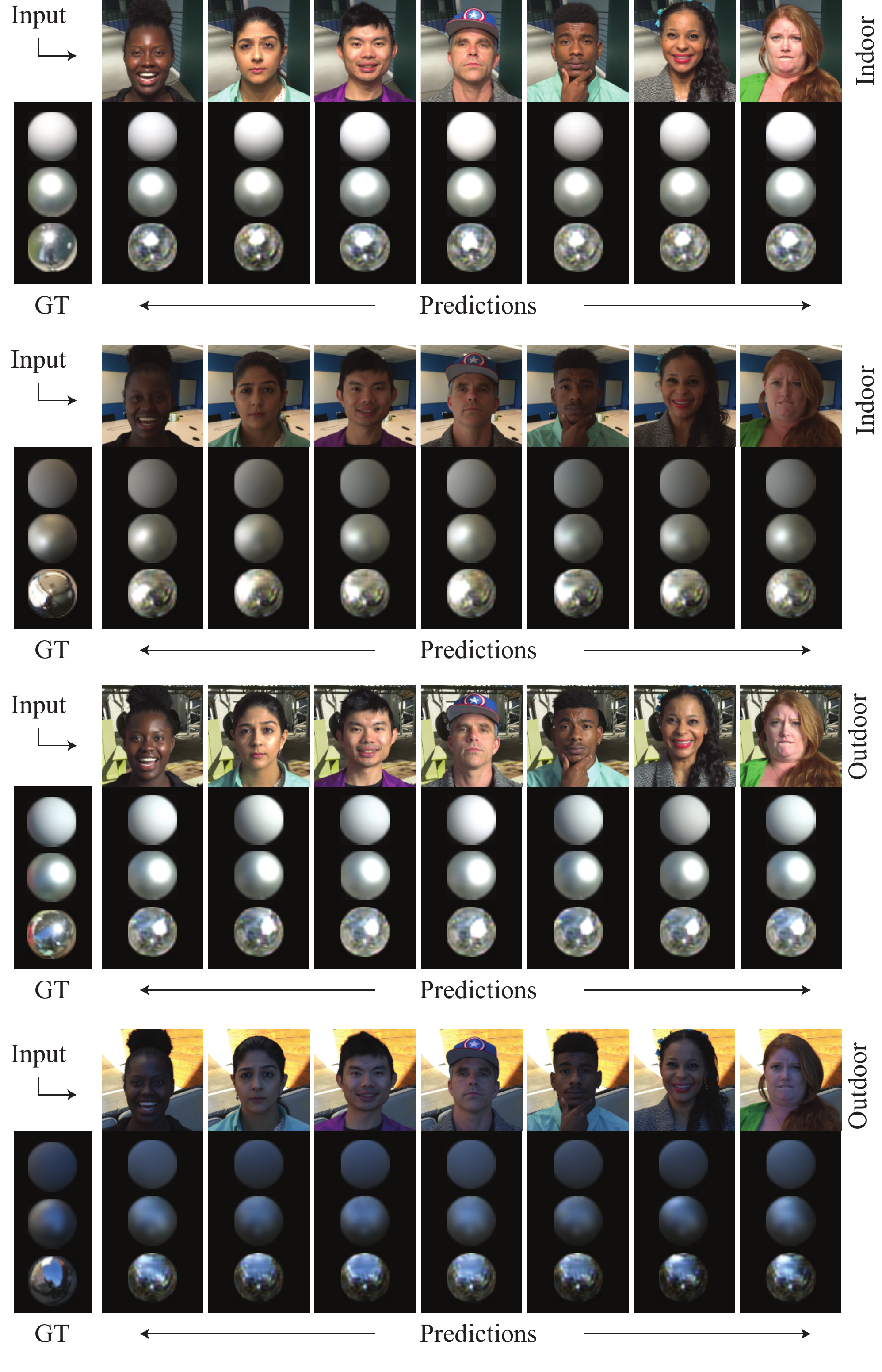}
\vspace{-20pt}
\caption{Left: spheres rendered with the ground truth illumination. Remaining columns: spheres rendered with the illumination produced using our technique, for input portraits of different subjects all lit with the same ground truth illumination. Our model recovers lighting at a similar scale for LDR input portraits of subjects with a variety of skin tones.}
\label{fig:ml_fairness_grid}
\vspace{-15pt}
\end{figure}

While $L_\text{rec}$ is a useful metric, its absolute value operation masks the sign of residual error. To see whether radiance is missing or added to the predicted lighting for each subject, we also show the total relative radiance difference [(GT-Pred.)/GT] for each color channel for each subject in Fig. \ref{fig:relative_radiance}. The trend lines in Fig. \ref{fig:relative_radiance} show that for evaluation subjects with smaller albedo values (measured as an average of each subject's forehead region), some energy in the estimated lighting is missing relative to the ground truth, with the inverse true for subjects with larger albedo values. For both indoor and outdoor scenes, this relative radiance difference is on average $\pm$20\% for evaluation subjects with very dark or very light skin tones, respectively, and smaller for subjects with medium skin tones. Nonetheless, as our evaluation subject with the lightest skin tone has an albedo value almost $3.5\times$ that of our evaluation subject with the darkest skin tone, the network has mostly learned the correct scale of illumination across diverse subjects. In Fig. \ref{fig:ml_fairness_grid}, we show examples where our model recovers similar lighting for different LDR input portraits of our evaluation subjects, where each is lit with the same ground truth illumination. In Fig. \ref{fig:relighting_grid}, we show that for a given input portrait (Fig. \ref{fig:relighting_grid}\textbf{a}), and lighting estimated from this portrait using our method Fig. \ref{fig:relighting_grid}\textbf{b}), we can accurately light a subject of a different skin tone (Fig. \ref{fig:relighting_grid}\textbf{f}) \textit{without} adjusting the scale of the illumination and composite them into the original image, closely matching that subject's ground truth appearance (Fig. \ref{fig:relighting_grid}\textbf{e}). An additional such example is shown in Fig. \ref{fig:teaser}.

\subsection{Lighting Consistency across Head Poses}
\label{sec:head_pose}

\begin{figure}[ht]
\vspace{-4pt}
\centering
\includegraphics[width=\linewidth]{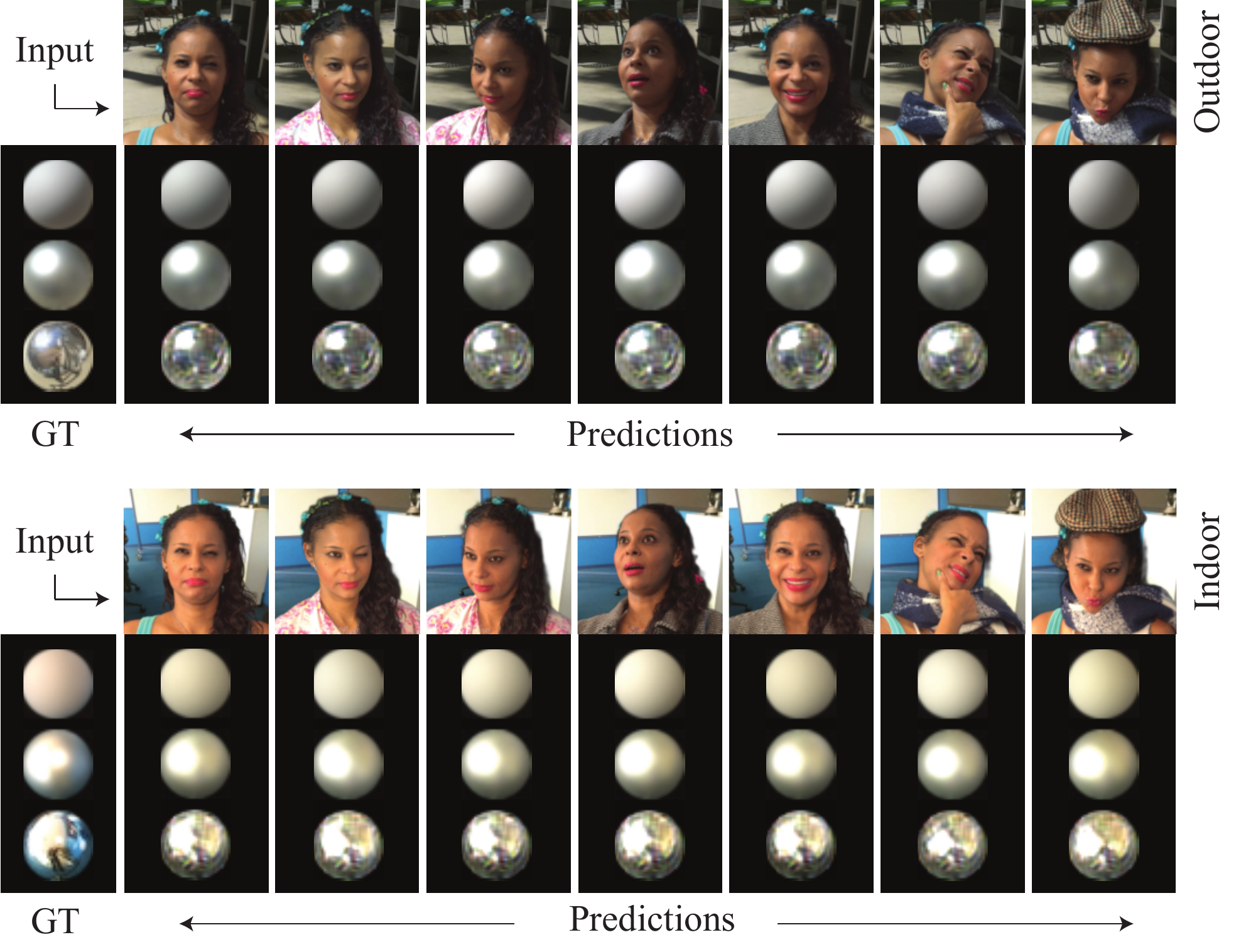}
\vspace{-15pt}
\caption{Left: spheres rendered with the ground truth illumination. Remaining columns: spheres rendered with the illumination produced using our technique, for input portraits of the same subject with different head poses and expressions, lit with the same illumination. Our method recovers similar lighting across facial expressions and head poses.}
\label{fig:head_pose}
\vspace{-3pt}
\end{figure}

We did not observe any marked differences in the lighting estimated for a given subject for different head poses or facial expressions. In Fig. \ref{fig:head_pose}, we show that similar illumination is recovered for different camera views and expressions for one of the evaluation subjects.

\subsection{Real-World Results}
\label{sec:real_world}

\begin{figure*}[t]
\centering
\includegraphics[width=\linewidth]{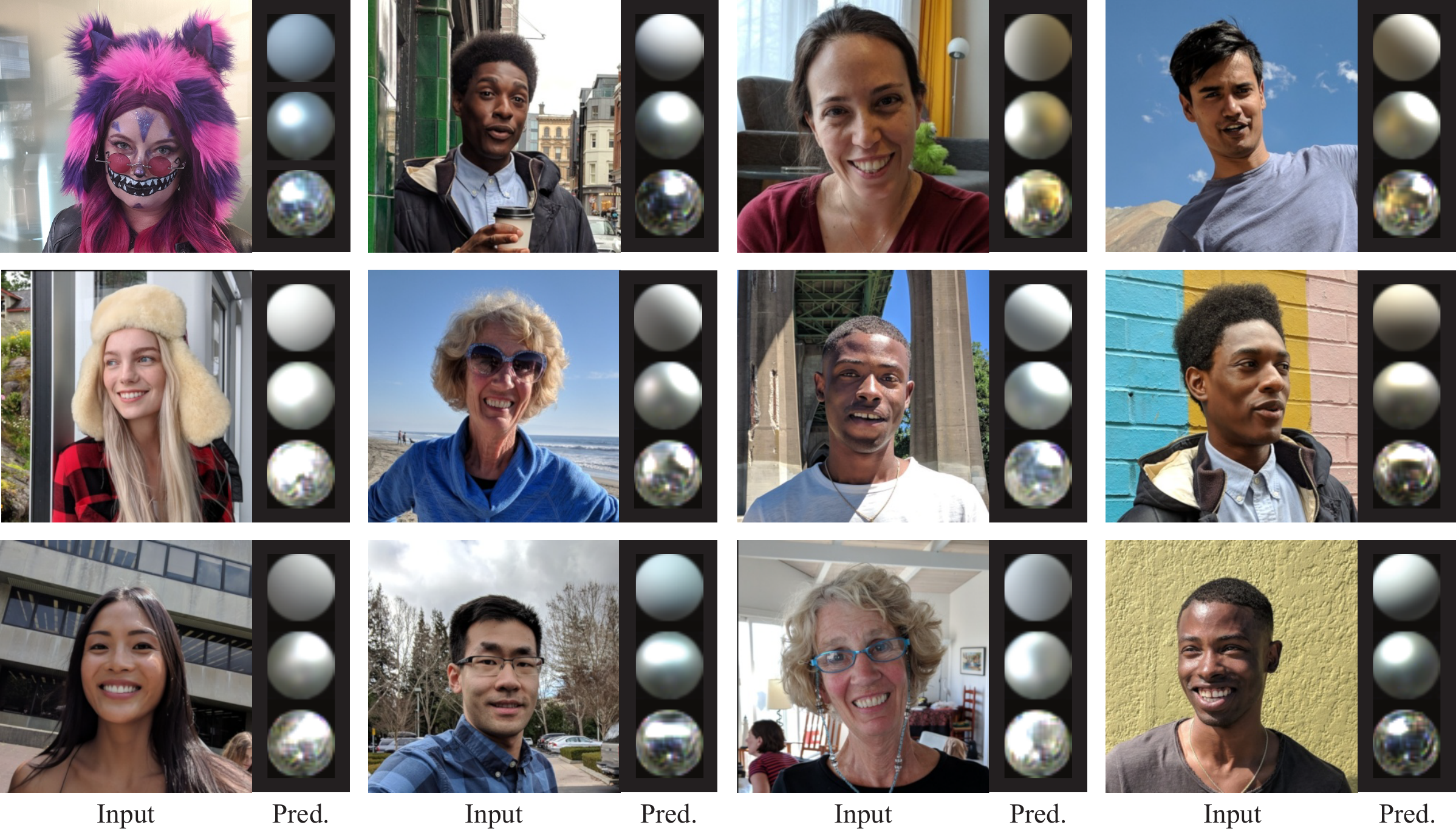}
\vspace{-20pt}
\caption{Diffuse, matte silver, and mirror spheres rendered using illumination estimated using our technique from the input portraits in-the-wild.}
\label{fig:wild}
\vspace{-15pt}
\end{figure*}

In Fig. \ref{fig:wild} we show lighting estimation from real-world portraits in-the-wild, for a diverse set of subjects, including one wearing a costume with face-paint. While ground truth illumination is not available, the sphere renderings produced using our lighting inference look qualitatively plausible. These results suggest that our model has generalized well to arbitrary portraits.
\section{Applications}
\label{sec:applications}
\paragraph{Mobile Augmented Reality} Our lighting inference runs in real-time on a mobile device (CPU: 27.5 fps, GPU: 94.3 fps on a Google Pixel 4 smartphone), enabling real-time rendering and compositing of virtual objects for smartphone AR applications. We show our inference running in real-time in our supplemental video. 

\begin{figure}[ht]
    \vspace{-3pt}
	\centerline{
		\begin{tabular}{@{}c@{ }c@{ }c@{}} 
			\includegraphics[height=1.9in]{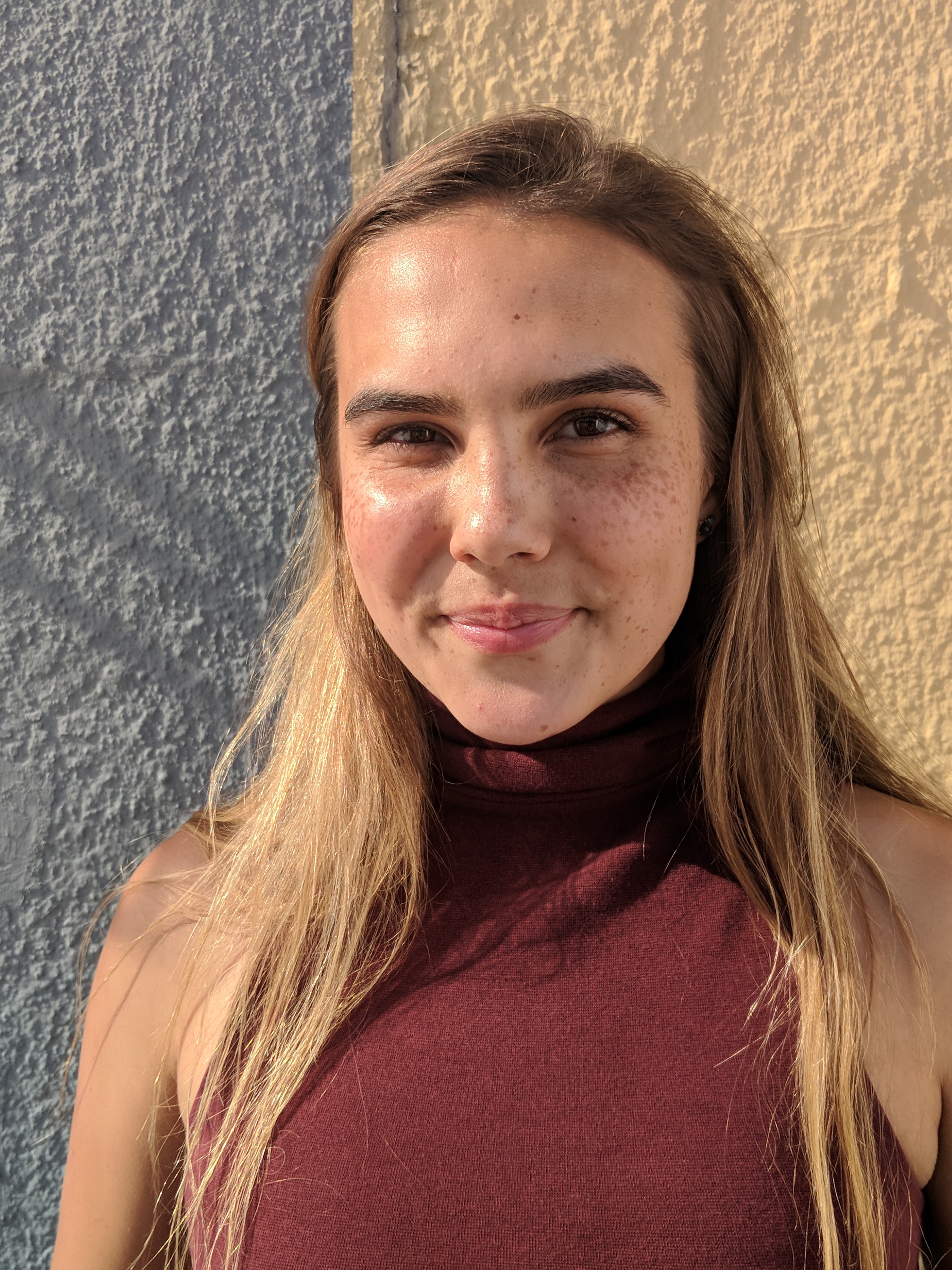} &
			\includegraphics[height=1.9in]{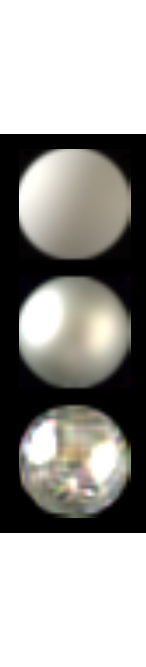} &
			\includegraphics[height=1.9in]{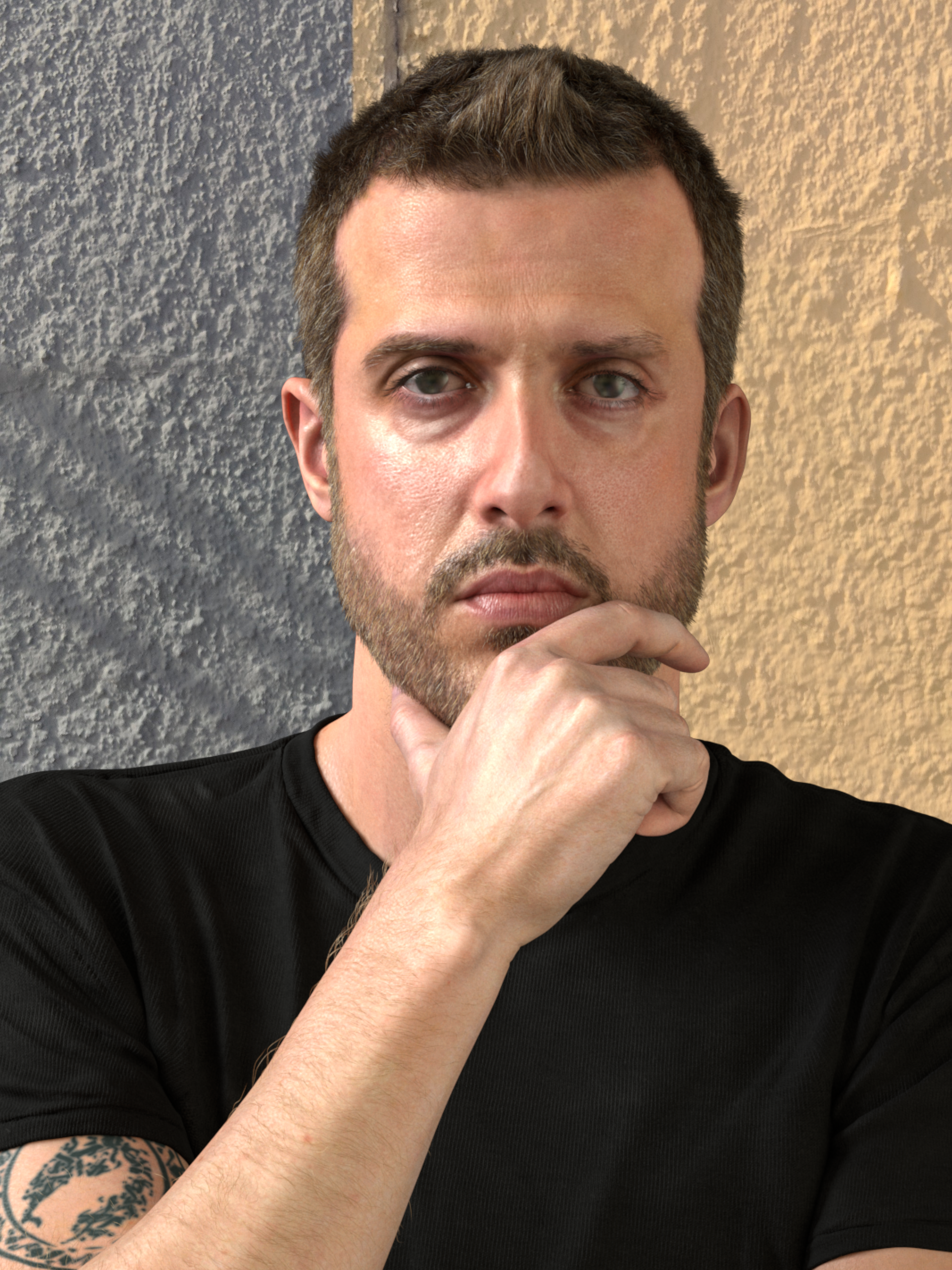} \\
			\includegraphics[height=1.9in]{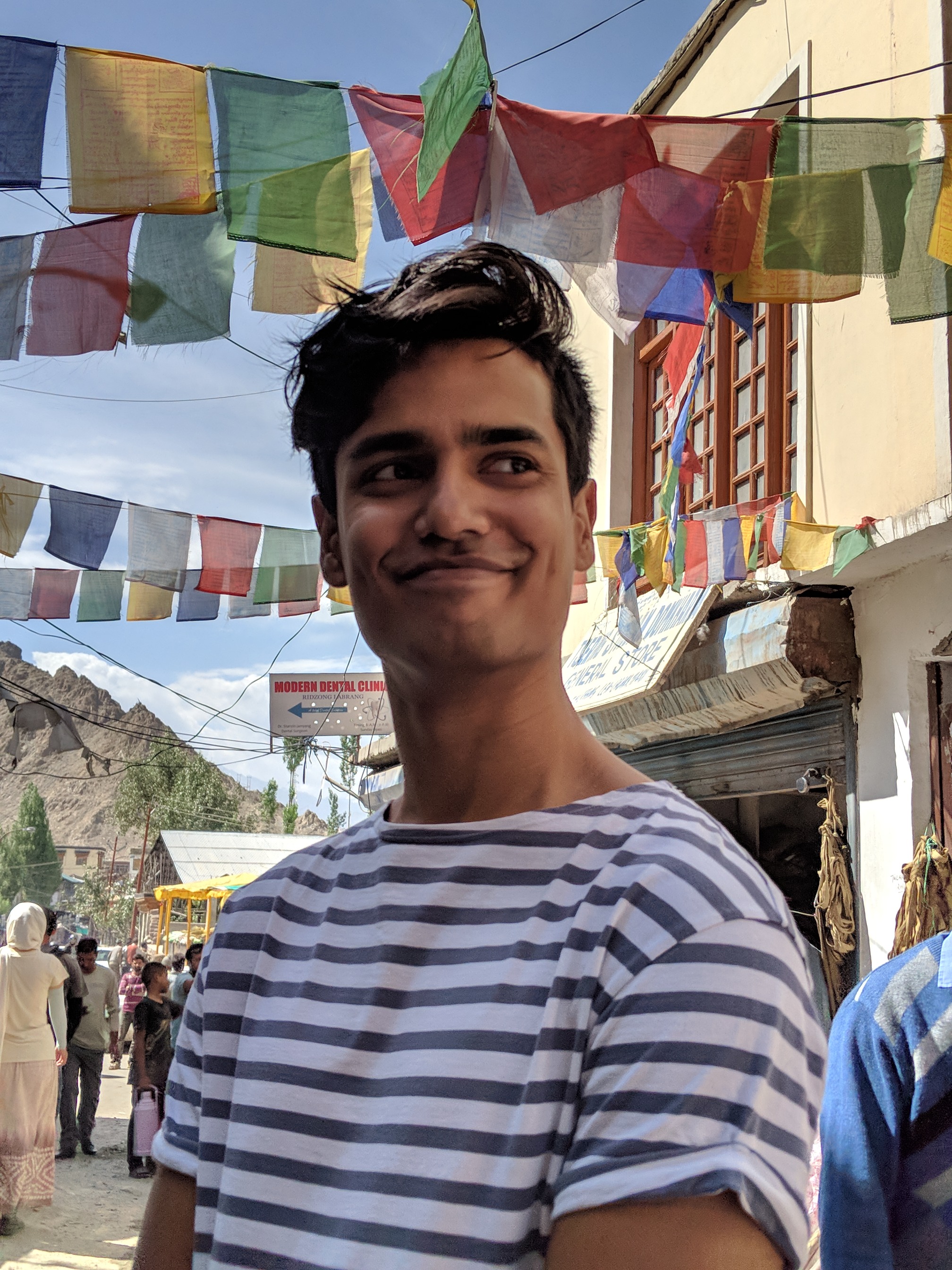} &
			\includegraphics[height=1.9in]{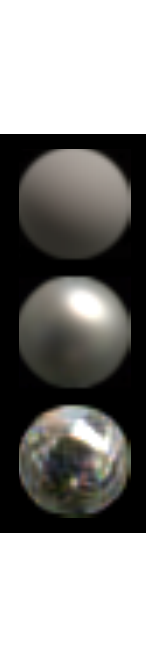} &
			\includegraphics[height=1.9in]{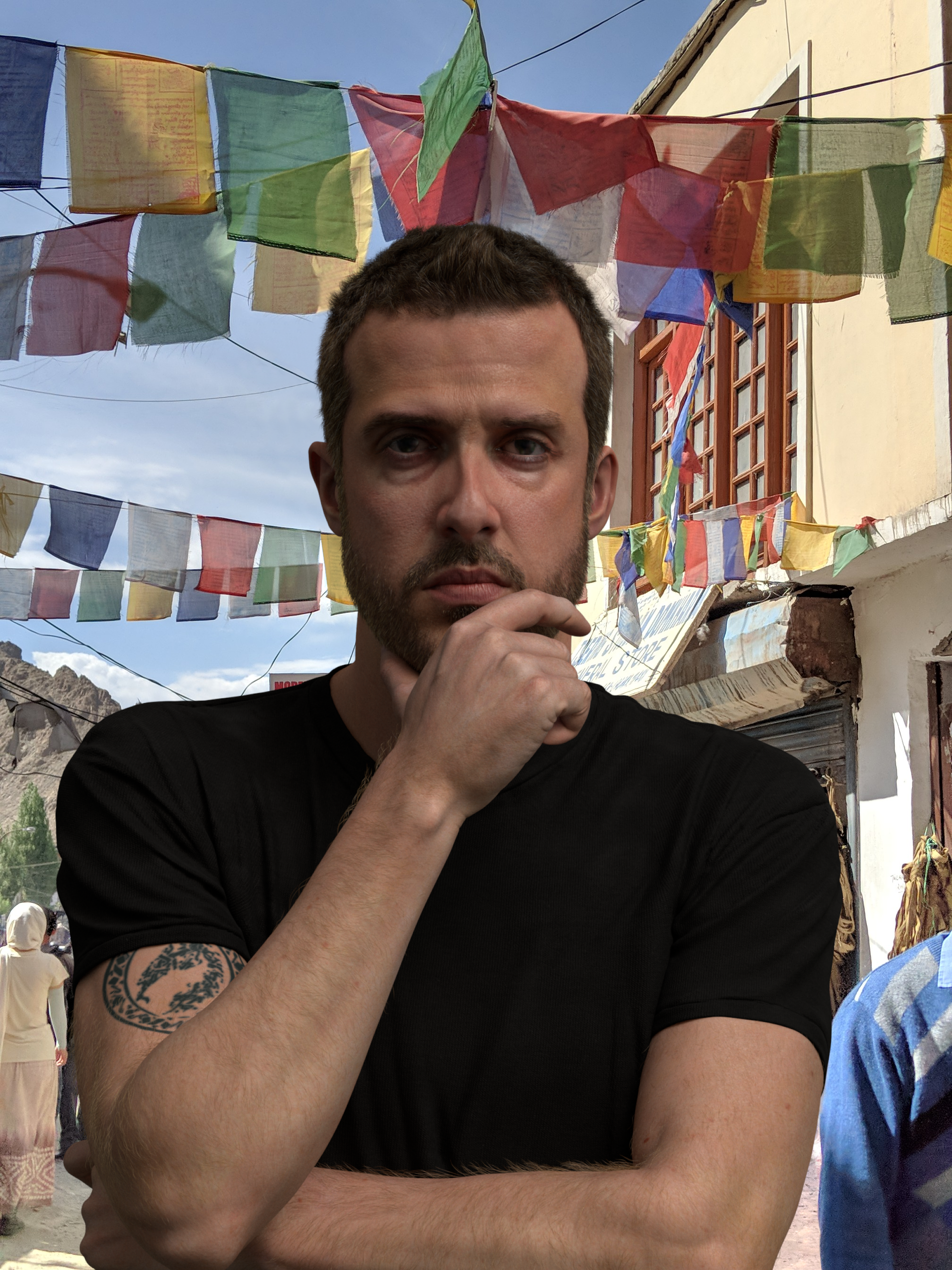} \\
			\footnotesize{\textbf{(a)}} input in-the-wild portrait & \footnotesize{\textbf{(b)} lighting} & \footnotesize{\textbf{(c)} digital character lit with \textbf{b}} \\
		\end{tabular}}
		\vspace{-8pt}
		\caption{\textbf{(a)} In-the-wild input portraits. \textbf{(b)} Lighting estimated by our technique. \textbf{(c)} A digital human character rendered with the predicted illumination, composited into the original scene. Digital character model by Ian Spriggs, rendered in V-Ray with the VRayAlSurfaceSkin shader.}
		\label{fig:ian_spriggs}
		\vspace{-8pt}
\end{figure}

\paragraph{Digital Double Actor Replacement} In Fig. \ref{fig:ian_spriggs}, we estimate lighting from in-the-wild portraits (\textbf{a}), and then light a virtual character to composite into the original scene with consistent illumination. These examples suggest that our method could be used for digital double actor replacement, without on-set lighting measurements.

\begin{figure}[ht]
    \vspace{-3pt}
	\centerline{
		\begin{tabular}{@{}c@{ }c@{}} 
			\includegraphics[width=1.4in]{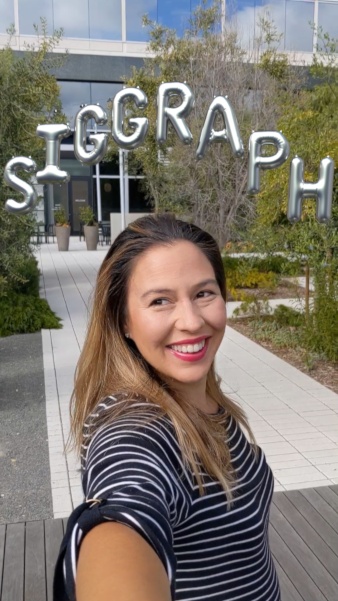} &
			\includegraphics[width=1.4in]{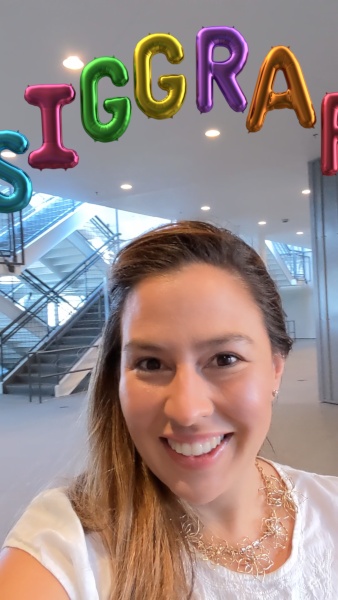}
			\\
		\end{tabular}}
		\vspace{-8pt}
		\caption{Virtual balloons composited into "selfie" portraits using lighting estimated by our technique.}
		\vspace{-5pt}
		\label{fig:virtual_object}
\end{figure}

\paragraph{Post-Production Virtual Object Compositing} In Fig. \ref{fig:virtual_object} we render and composite a set of shiny virtual balloons into a "selfie" portrait, using lighting estimates produced by our method. We show a version with motion in our supplemental video. 

\section{Limitations}

\begin{figure}[h]
\vspace{-5pt}
\centering
\includegraphics[width=2.7in]{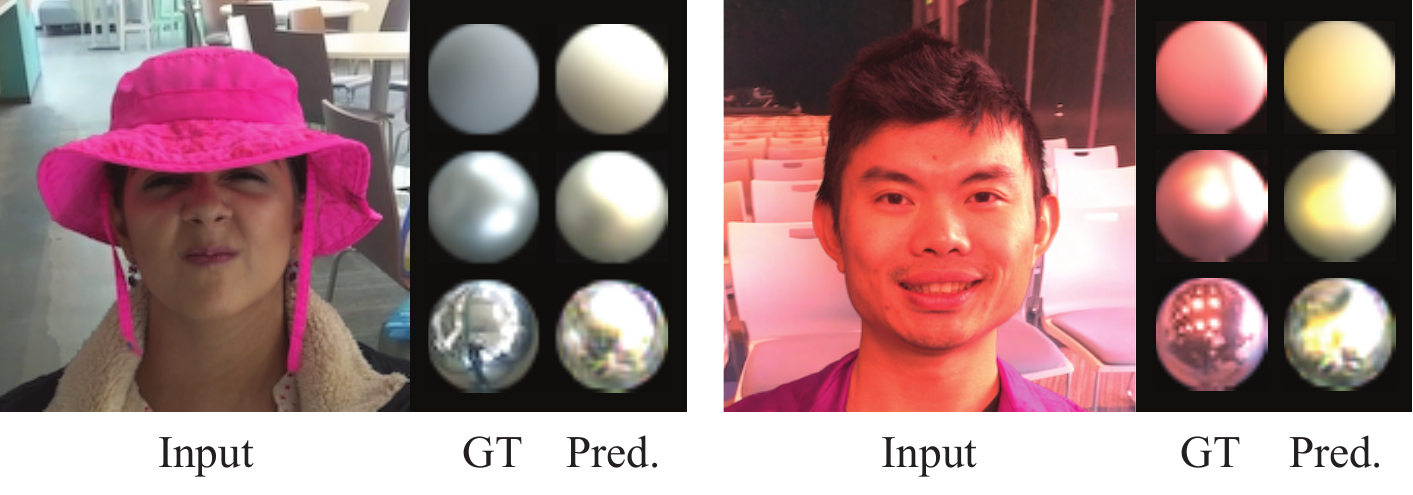}
\vspace{-10pt}
\caption{Two example failure cases. Left: A brightly-hued hat not observed in training. Right: Input lighting environment with non-natural color balance.}
\label{fig:failures}
\vspace{-10pt}
\end{figure}

As our method relies on a face detector, it fails if no face is detected. Fig. \ref{fig:failures} shows two other failure modes: an example where a saturated pink hat not observed in training leads to an erroneous lighting estimate, and an example where the illumination color is incorrectly estimated for an input environment with unnatural color balance. This input example was generated by scaling the red channel of the ground truth illumination by a factor of 3. Future work could address the first limitation with additional training data spanning a broader range of accessories, while the second limitation could be addressed with data augmentation via adjusting the white balance of the ground truth illumination. Finally, our lighting model assumes distant illumination, so our method is not able to recover complex local lighting effects.

\section{Conclusion}

We have presented a learning-based technique for estimating omnidirectional HDR illumination from a single LDR portrait image. Our model was trained using a photo-realistic, synthetically-rendered dataset of portraits with ground truth illumination generated using reflectance fields captured in a light stage, along with more than one million lighting environments captured using an LDR video-rate technique, which we promoted to HDR using a novel linear solver formulation. We showed that our method out-performs both the previous state-of-the-art in portrait-based lighting estimation, and, for non-Lambertian materials, a low-frequency, second order spherical harmonics decomposition of the ground truth illumination. We are also, to the best of our knowledge, the first to explicitly evaluate our lighting estimation technique for subjects of diverse skin tones, while demonstrating recovery of a similar scale of illumination for different subjects. Our technique runs in real-time on a mobile device, suggesting its usefulness for improving the photo-realism of face-based augmented reality applications. We further demonstrated our method's utility for post-production visual effects, showing that digital characters can be composited into real-world photographs with consistent illumination learned by our model. 

\bibliographystyle{ACM-Reference-Format}
\bibliography{bibliography}


\end{document}